\documentclass[11pt]{article}

\usepackage[preprint]{acl}

\usepackage{times}
\usepackage{latexsym}

\usepackage[T1]{fontenc}

\usepackage[utf8]{inputenc}

\usepackage{placeins}
\usepackage{framed}

\usepackage{microtype}

\usepackage{inconsolata}

\usepackage{graphicx}

\newcommand\todo[1]{{\color{black}{#1}}}%
\newcommand\rev[1]{{\color{black}{#1}}}%
\newcommand\final[1]{{\color{black}{#1}}}%
\newcommand\ok[1]{{\color{black}{#1}}}%
\usepackage{amsmath}   
\usepackage{booktabs}  
\usepackage{enumitem}  
\usepackage{tabularx}  

\newcommand{\abNModelsWord}{ten}

\newcommand{\abNJudgesWord}{three}

\newcommand{\abNConditionsWord}{six}
\newcommand{\abNDims}{10}
\newcommand{\abNDimsWord}{ten}
\newcommand{\abNDimsMinusOneWord}{nine}

\newcommand{\abNParaphrasesWord}{five}
\newcommand{\abWardMax}{10}
\newcommand{\abNConstrainedWord}{three}

\newcommand{\abNPerCell}{30}
\newcommand{\abNPerModel}{180}
\newcommand{\abNGenerations}{1{,}800}

\newcommand{\abTemperature}{1.0}
\newcommand{\abMaxTokens}{64{,}000}

\newcommand{\abMaxTokensLlamaMaverick}{16{,}384}

\newcommand{\abMaxTokensModelLlamaMaverick}{Llama 4 Maverick}

\newcommand{\abJudgeTemperature}{0}
\newcommand{\abMaxParseAttemptsWord}{three}
\newcommand{\abJudgeMaxTokens}{8{,}000}
\newcommand{\abJudgeMaxTokensCeiling}{32{,}000}

\newcommand{\abBootstrapB}{10{,}000}
\newcommand{\abCILevel}{95}

\newcommand{\abTokensPerRateUnit}{100}
\newcommand{\abFeatureMaxWords}{15}
\newcommand{\abAlphaGood}{0.60}
\newcommand{\abScoreEarth}{0}
\newcommand{\abScoreDeparture}{1}

\newcommand{\abExampleLegsWord}{six}

\newcommand{\abPowerTarget}{80}
\newcommand{\abJudgePanel}{Claude Opus 4.6 (2026-02-04), Gemini 3.1 Pro (2025-12-17), and GPT-5.4 (2026-03-05)}
\newcommand{\abJudgeNameClaudeOpus}{Claude Opus 4.6}

\newcommand{\abJudgeNameGPT}{GPT-5.4}

\newcommand{\abGravityMultiplier}{10}
\newcommand{\abBandExcellent}{0.80}
\newcommand{\abBandGood}{0.60}
\newcommand{\abBandModerate}{0.40}
\newcommand{\abBandFair}{0.20}
\newcommand{\abHumanN}{100}
\newcommand{\abHumanPlatform}{Prolific}
\newcommand{\abHumanApproval}{95}
\newcommand{\abHumanPriorTasks}{100}
\newcommand{\abHumanPay}{\pounds{}6.13}
\newcommand{\abHumanMinutes}{11}
\newcommand{\abHumanSeconds}{10}
\newcommand{\abHumanExcludedWord}{two}

\newcommand{\abHumanEstMinutes}{5}
\newcommand{\abHumanReward}{\pounds{}0.75}
\newcommand{\abHumanAdvertisedRate}{\pounds{}9.00}

\newcommand{\abHumanPayPerParticipant}{\pounds{}1.14}

\newcommand{\abHumanBioNone}{5}

\newcommand{\abHumanBioSchool}{49}

\newcommand{\abHumanBioUndergrad}{27}

\newcommand{\abHumanBioGradNCap}{Nineteen}

\newcommand{\abPythonVersion}{3.12.3}
\newcommand{\abNumpyVersion}{2.4.6}
\newcommand{\abScipyVersion}{1.17.1}
\newcommand{\abPandasVersion}{2.3.1}
\newcommand{\abKrippendorffVersion}{0.8.2}
\newcommand{\abTokensInputM}{15.3}
\newcommand{\abTokensOutputM}{8.7}

\newcommand{\abTokensReasoningM}{2.0}
\newcommand{\abTokensParaphraseM}{6.0}
\newcommand{\abNParaGenCalls}{500}
\newcommand{\abNParaJudgeCalls}{1{,}500}
\newcommand{\abNHumanJudgeCalls}{300}

\newcommand{\abHoursGen}{34.5}
\newcommand{\abHoursJudge}{31.2}
\newcommand{\abJudgesViaOpenRouter}{Claude Opus 4.6 and GPT-5.4}

\newcommand{\abBestModelSized}{Ministral 3 14B}
\newcommand{\abBestMean}{6.55}
\newcommand{\abBestMeanCI}{6.55 [6.19, 6.89]}
\newcommand{\abWorstModel}{GPT-6 Luna}
\newcommand{\abWorstModelSized}{GPT-6 Luna}
\newcommand{\abWorstMean}{3.28}
\newcommand{\abWorstMeanCI}{3.28 [2.87, 3.70]}
\newcommand{\abSecondModels}{Grok 4.7}

\newcommand{\abLeadMargin}{1.1}
\newcommand{\abBestModelShort}{Ministral}

\newcommand{\abNOpenWord}{five}
\newcommand{\abOpenMean}{5.40}

\newcommand{\abNClosedWord}{five}
\newcommand{\abClosedMean}{4.58}
\newcommand{\abOpenClosedDiff}{0.82}
\newcommand{\abOpenClosedDiffCI}{[-0.01, 1.75]}
\newcommand{\abOpenClosedPermP}{$p = 0.135$}
\newcommand{\abOpenClosedNSplits}{252}

\newcommand{\abBestClosedModelSized}{Grok 4.7}
\newcommand{\abBestClosedMean}{5.42}

\newcommand{\abWorstClosedModelSized}{GPT-6 Luna}
\newcommand{\abWorstClosedMean}{3.28}

\newcommand{\abOpenAboveClosedModelsSized}{Ministral 3 14B}

\newcommand{\abCondMinBaseline}{2.16}
\newcommand{\abCondMaxBaseline}{6.88}

\newcommand{\abCondMinDeparturePrimed}{8.68}

\newcommand{\abNCondLeadersWord}{four}

\newcommand{\abNCondLeadersOpenWord}{two}

\newcommand{\abDensestModelSized}{Grok 4.7}

\newcommand{\abDensestRate}{0.60}
\newcommand{\abDensestTokens}{968}

\newcommand{\abSparsestModelShort}{DeepSeek}
\newcommand{\abSparsestRate}{0.09}

\newcommand{\abBestModelTokens}{1{,}916}
\newcommand{\abMaxCapSharePct}{26}
\newcommand{\abKruskalH}{972.96}
\newcommand{\abKruskalP}{$p < .001$}
\newcommand{\abRConstrainedAmmonia}{0.37}
\newcommand{\abRCIConstrainedAmmonia}{$[0.28, 0.45]$}
\newcommand{\abPConstrainedAmmonia}{$p < .001$}
\newcommand{\abRConstrainedHighGravity}{-0.53}
\newcommand{\abRCIConstrainedHighGravity}{$[-0.61, -0.46]$}
\newcommand{\abPConstrainedHighGravity}{$p < .001$}
\newcommand{\abRConstrainedNoLight}{0.01}
\newcommand{\abRCIConstrainedNoLight}{$[-0.08, 0.10]$}
\newcommand{\abPConstrainedNoLight}{$p = 1.0$}
\newcommand{\abRDeparturePrimed}{0.97}
\newcommand{\abRCIDeparturePrimed}{$[0.95, 0.99]$}
\newcommand{\abPDeparturePrimed}{$p < .001$}
\newcommand{\abRDetailedDescription}{0.23}
\newcommand{\abRCIDetailedDescription}{$[0.14, 0.32]$}
\newcommand{\abPDetailedDescription}{$p < .001$}

\newcommand{\abBonferroniAlpha}{0.003}
\newcommand{\abMinDetectableR}{0.20}

\newcommand{\abHumanMean}{1.93}
\newcommand{\abHumanMeanCI}{[1.57, 2.31]}

\newcommand{\abAlphaWard}{0.913}
\newcommand{\abAlphaWardBand}{Excellent}

\newcommand{\abNGoodDimsWord}{three}

\newcommand{\abNModerateDimsWord}{one}

\newcommand{\abNClearDimsWord}{nine}
\newcommand{\abBelowGoodDims}{sensory organs}
\newcommand{\abBelowGoodMin}{0.422}
\newcommand{\abNExtractionCalls}{5{,}400}

\newcommand{\abParaN}{10}
\newcommand{\abParaNPairs}{30}
\newcommand{\abParaNUniquePairs}{15}
\newcommand{\abParaNPermutations}{10{,}000}
\newcommand{\abParaRhoBar}{0.9}
\newcommand{\abParaMadBar}{1.0}
\newcommand{\abParaMadBarPct}{10}
\newcommand{\abParaRhoMin}{0.58}
\newcommand{\abParaRhoMax}{0.97}
\newcommand{\abParaRhoMedian}{0.75}
\newcommand{\abParaNAboveNull}{23}
\newcommand{\abParaNullHiMin}{0.63}
\newcommand{\abParaNullHiMax}{0.66}
\newcommand{\abParaMadMin}{0.49}
\newcommand{\abParaMadMax}{2.38}
\newcommand{\abParaMadMedian}{1.21}

\newcommand{\abParaNRhoBarWord}{two}

\newcommand{\abParaNUniqueRhoBarWord}{one}

\newcommand{\abParaNUniqueMadBarWord}{six}

\newcommand{\abParaNUniqueBothBarsWord}{one}
\newcommand{\abParaOutlierModel}{Mistral Large 4}

\newcommand{\abParaOutlierBase}{5.16}
\newcommand{\abParaOutlierMin}{5.27}
\newcommand{\abParaOutlierMax}{7.97}

\newcommand{\abDimRhoBar}{0.9}
\newcommand{\abDimTauBar}{0.8}
\newcommand{\abDimLooRhoMin}{0.93}
\newcommand{\abDimLooRhoMax}{1.00}

\newcommand{\abDimMadMin}{0.17}
\newcommand{\abDimMadMinDim}{habitat}
\newcommand{\abDimMadMax}{0.87}
\newcommand{\abDimMadMaxDim}{sensory organs}

\newcommand{\abDimMinKRho}{1}

\newcommand{\abDimMinKRhoPValue}{0.002}
\newcommand{\abDimMinKTau}{1}

\newcommand{\abDimMinKTauPValue}{0.001}

\newcommand{\abDimMadBelowOneK}{7}

\newcommand{\abDimBestSingleDim}{Metabolism / Energy Source}
\newcommand{\abDimBestSingleRho}{0.95}
\newcommand{\abDimSubsetMedianRhoOne}{0.49}
\newcommand{\abDimSubsetMedianRhoBarK}{6}
\newcommand{\abDimFeatureMeanRho}{0.34}
\newcommand{\abJudgeNPermutations}{10{,}000}
\newcommand{\abJudgeRhoBar}{0.9}
\newcommand{\abJudgeAlphaTentative}{0.667}
\newcommand{\abJudgeAlphaDropBar}{0.1}

\newcommand{\abJudgeSingleRhoMin}{0.89}
\newcommand{\abJudgeSingleRhoMax}{0.99}

\newcommand{\abJudgeSingleNClearBarCap}{Two}

\newcommand{\abJudgeMissName}{Claude Opus 4.6}
\newcommand{\abJudgeMissRho}{0.89}
\newcommand{\abJudgeLooAlphaMin}{0.87}
\newcommand{\abJudgeLooAlphaMax}{0.94}

\newcommand{\abJudgeLooAlphaClaudeOpus}{0.94}

\newcommand{\abJudgeLooAlphaGPT}{0.87}

\newcommand{\abSfNZeroWord}{three}

\newcommand{\abSfZeroModels}{Gemini 3.8 Flash, GPT-6 Astra, and GPT-6 Luna}

\newcommand{\abSfNNegativeWord}{one}

\newcommand{\abSfNegativeModels}{Claude Opus 5.5}

\newcommand{\abHumanNOffensiveCap}{One}

\newcommand{\abHumanPriorYesCap}{Twenty}

\newcommand{\abNExcellentDimsCap}{Six}

\newcommand{\abSfNPairsCap}{Four}

\newcommand{\abDimNRankPreservedCap}{Two}

\newcommand{\abHumanFemale}{70}
\newcommand{\abHumanMale}{30}
\newcommand{\abHumanAgeMedian}{29}
\newcommand{\abHumanAgeMin}{18}
\newcommand{\abHumanAgeMax}{69}
\newcommand{\abHumanCountryOne}{South Africa}

\newcommand{\abHumanCountryOneNCap}{Fifty-eight}
\newcommand{\abHumanCountryTwo}{United Kingdom}
\newcommand{\abHumanCountryTwoN}{27}

\newcommand{\abHumanCountryRestN}{15}
\newcommand{\abHumanCountryRestWord}{six}

\title{AlienBench: A Benchmark of Structural Novelty in \\ Language Model Outputs}

\author{Jonas Oppenlaender\\
Centre for Applied Computing\\
University of Oulu\\
Oulu, Finland \\
  \texttt{jonas.oppenlaender@oulu.fi} \\\And
Johanna Silvennoinen \\
Faculty of Information Technology  \\
University of Jyv\"askyl\"a \\
Jyv\"askyl\"a, Finland \\
  \texttt{johanna.silvennoinen@jyu.fi} \\}

\begin{document}
\maketitle
\begin{abstract}
Language models are increasingly used for open-ended creative tasks.
In these uses, it matters whether a model's ideas depart from familiar examples or stay close to them.
The structured imagination task from cognitive psychology provides a fixed reference for this departure.
In this task, people imagine a creature from another planet, and departures from features of Earth animals measure creativity.
We introduce \textbf{AlienBench}, a benchmark that adapts this task to language models.
Models describe alien creatures under \abNConditionsWord{} prompt conditions that vary the task framing and the planetary environment.
A panel of \abNJudgesWord{} LLM judges labels \abNDimsWord{} biological features of each description as Earth-typical or as a departure.
The number of departures is the Ward Departure Score (0--\abWardMax{}).
The judges agree closely (Krippendorff's $\alpha = \abAlphaWard{}$).
Across \abNModelsWord{} proprietary and open-weight models, overall scores range from \abWorstMean{} (\abWorstModel{}) to \abBestMean{}
(\abBestModelSized{}).
No single model scores highest under every condition.
Under the baseline prompt, every model has a higher mean score than human participants ($N = \abHumanN{}$).
AlienBench thus provides a feature-level measure of \textit{structural novelty}, the departure in kind from familiar examples.
\end{abstract}


\section{Introduction}
\label{sec:introduction}

\final{%
	Language models are increasingly used for open-ended creative tasks, such as writing \citep{10.1073/pnas.2504966122}, design \citep{10.1145/3661455.3669868}, and idea generation \citep{3613904.3642414}.
}%
\ok{%
	In these tasks, a model is often asked to produce something original.
	The model's output may depart from familiar examples or stay close to them.
	Telling these two cases apart requires a measure of how far an output departs from familiar examples.
}%

\ok{%
	Existing creativity evaluations do not provide such a measure.
Creativity is often measured with a single holistic score for each output.
In the Consensual Assessment Technique \citep[CAT;][]{amabile1982cat}, expert raters assign such scores to creative products.
LLM judges can also assign holistic scores, as in the single-answer grading of MT-Bench \citep{zheng2023mtbench}.
However, a holistic score does not show which features of an output depart from familiar examples.
This matters wherever language models are used to generate novel ideas.
A model asked to imagine the unknown may return a version of the known.
A model that stays close to familiar patterns may narrow the range of possibilities that we consider divergent \citep{jiang2026artificial}.
}%

\final{%
	This paper focuses on one component of creativity, which we call \textit{structural novelty}.
	An output is structurally novel on a feature if the feature differs in kind from the corresponding feature of familiar examples.
	Differences in size, number, or appearance alone do not count as structural novelty.
	For instance, a dog with five legs is not structurally novel, because it changes only the number of a familiar feature (i.e., the dog's four legs).
}

\final{%
Measuring structural novelty requires a task with four properties.
First, the task asks for a new member of an existing category, such as a new animal.
Second, the category has known default features to which the features of the new member can be
compared, such as legs for locomotion.
Third, for each feature, a decision rule specifies which descriptions match the default and which depart from it.
Under such a rule, a dog with five legs still moves on legs, whereas a creature that rolls does not.
Fourth, the task describes these features, so that the feature's departure from the default can be scored separately.%
}%

\final{%
The structured imagination task of \citet{ward1994} meets all four requirements.
In this task, participants imagine and describe a creature that lives on a planet other than Earth.
\citet{ward1994} found that in human participants, most descriptions of these creatures share basic features with Earth animals, such as bilateral symmetry, legs, and a central nervous system.
Ward attributed this result to prior knowledge.
When people generate a new instance of a category (e.g., an alien creature), they reuse features of the instances they have already observed on Earth (e.g., an animal from Earth).
In this task, Earth animals are the familiar examples, and their typical features are the default features of the category, which we call
\textit{Earth-typical defaults}.
}%

\final{%
We introduce \textbf{AlienBench}, a benchmark that adapts the structured imagination task to language models.
Each model describes an alien creature under \abNConditionsWord{} prompt conditions.
The conditions differ in how the task is framed and in the physical environment of the planet.
A panel of \abNJudgesWord{} LLM judges labels \abNDimsWord{} biological features in each LLM-generated description, such as locomotion and body symmetry, as Earth-typical (\abScoreEarth{}) or as a departure (\abScoreDeparture{}).
A feature counts as a departure only if it differs in kind from the Earth default.
For example, a creature with \abExampleLegsWord{} legs is Earth-typical for locomotion because it still walks on legs.
A creature that rolls or moves by jet propulsion is labeled as a departure.
Absence of a feature in the description is scored as 0.
The sum of the \abNDimsWord{} labels is the \textit{Ward Departure Score} (WDS), which ranges from 0 to \abWardMax{}.
Each LLM judge assigns this score to every description, and the scores of the \abNJudgesWord{} judges are averaged per description.
In AlienBench, each model generates \abNPerCell{} descriptions per prompt condition, 
sampled at temperature \abTemperature{} with a different random seed.
A model's overall score is the mean Ward Departure Score over its \abNPerModel{} descriptions, with equal weight to each prompt condition.
}%

\final{We evaluate \abNModelsWord{} LLMs (\abNGenerations{} generations in total).
The \abNJudgesWord{} LLM judges agree closely on the Ward Departure Score of each description (Krippendorff's $\alpha = \abAlphaWard{}$).}
\ok{%
%
The lowest overall score comes from a proprietary model (\abWorstMean{} for \abWorstModel{}).
\abBestModelSized{} has the highest overall score among the tested models.
%
Among the proprietary models, \abBestClosedModelSized{} performs best (\abBestClosedMean{}).
We also collected alien creature descriptions from \abHumanN{} human participants under the baseline prompt, and find that all models outperform this human sample in the baseline condition.
}%

\final{%
The benchmark conflates two dimensions of divergent thinking: \textit{originality}, which the benchmark measures as structural novelty, and \textit{elaboration}, approximated by output length.
This conflation corresponds to a realistic condition of text production in real-world settings.
For instance, human writers decide how much time and effort to invest in a task (i.e., elaboration), and the resulting text depends on this decision as well as on the 
originality of their ideas.
Language models differ in the amount of text they produce for the same prompt.
Thus, the benchmark evaluates generated text as it is delivered in practice and not \todo{creative ability} in isolation.
}%

\ok{%
This paper makes the following contributions:%
}%
\begin{itemize}%
\item%
\ok{%
	AlienBench, a benchmark that adapts the structured imagination task of \citet{ward1994} to language models.
	AlienBench consists of \abNDimsWord{} features with explicit rules for how to score a binary departure, \abNConditionsWord{} prompt conditions, and an overall model score that averages the Ward Departure Score over repeated generations from all \abNConditionsWord{} conditions.
	A panel of \abNJudgesWord{} LLM judges applies these rules with high agreement (Krippendorff's $\alpha = \abAlphaWard{}$).
}%
\item
\ok{%
	An evaluation of structural novelty in the outputs of \abNClosedWord{} proprietary and \abNOpenWord{} open-weight language models.
}%
\ok{%
	Overall scores range from \abWorstMean{} to \abBestMean{} (on a 1--10 scale).
}%
\item
\ok{%
	A human baseline from 
	human participants ($N = \abHumanN{}$) who completed the task under the same baseline prompt as the language models.
	All language models achieve higher scores than the human participants in the baseline condition.
}%
\end{itemize}%
\final{%
	We release the code, 
	generations, LLM judge outputs, human baseline data, and results at \href{https://huggingface.co/datasets/AlienBench/alienbench}{https://huggingface.co/datasets/AlienBench/alien bench} (code: MIT, data: CC-BY 4.0; Appendix~\ref{app:datacard}).%
}%

\section{Related Work}%
\label{sec:relatedwork}%
%
\final{%
	Creativity is commonly defined as the production of novel and useful outputs \citep{runco2012standarddef}.
	Psychometric measures of creativity operationalize this definition with divergent-thinking tasks, such as the Alternative Uses Task (AUT) \citep{guilford1967} and the Divergent Association Task (DAT) \citep{olson2021dat}, or with expert ratings of creative products under the Consensual Assessment Technique (CAT) \citep{amabile1982cat}.
}%

\final{%
Research on the creativity of language models follows three lines.
	The first line 
	applies psychometric divergent-thinking tasks to language models.
	\citet{koivisto2023aut} administered the AUT to language models, and \citet{chen2023dsa} used the DAT and semantic distance as a proxy for creativity.
}%
\final{%
	The second line introduces 
		new benchmarks designed for language models.
	For instance,
	\citet{tian2024macgyver} introduced MacGyver, a benchmark of unconventional-use problems under physical constraints, and \citet{si2025researchideas} evaluated whether model-generated research ideas are judged novel by domain experts.
}%
\final{%
	The third line scores novelty by comparing a model's outputs with a reference, such as existing text or the model's other outputs.
	For instance, 
	the Creativity Index \citep{lu2025creativityindex} measures how much of a model's text can be reconstructed from snippets of existing text on the Web.
	NoveltyBench \citep{zhang2025noveltybench} measures the diversity of a model's outputs across repeated samples for the same prompt.%
}%

\ok{%
	These lines of work score different units of output:
	whole outputs by expert ratings, the words of a response by their semantic distance, the spans of a text by their overlap with existing text, or a set of outputs by their diversity.
%
%
	In contrast, AlienBench evaluates the features of an output against fixed Earth-typical defaults.
}%


\final{%
	The creative cognition framework of \citet{finke1992creativecognition} argued for constrained generation tasks whose outputs show which prior knowledge people use.
	The structured imagination task of \citet{ward1994} is an instance of such a task.
	In this task, participants imagine a novel exemplar of a known category.
	The number of features that depart from the category's defaults measures how closely a new instance follows prior knowledge.
	Participants who approach the task at an abstract level, rather than starting from a specific familiar instance of the category, produce descriptions that are rated as more original.
	The task has since been used in further studies of creative cognition with human participants \citep{kozbeltDurmysheva2007,wardWickes2009}.
	AlienBench adapts the task of \citet{ward1994} to language models.
}%

\ok{%
	A separate line of work concerns the reliability of LLM judges.
	\citet{zheng2023mtbench} studied LLM-as-a-judge evaluation with MT-Bench.
	Concurrent work used LLMs to evaluate generated summaries and dialogue responses \citep{liu2023geval} and to simulate human preference annotations
	\citep{dubois2023alpacafarm}.
	These studies report agreement between strong LLM judges and human annotators on open-ended tasks.
	Dedicated evaluators such as Prometheus \citep{kim2024prometheus} train open-source models as judges.
	Subsequent work characterized systematic biases in LLM judges: \citet{wang2024faireval} documented position and verbosity bias, and \citet{panickssery2024selfpreference} documented self-preference bias, the tendency of a judge to assign more favorable scores to its own generations.
	A panel of judges from different model families	avoids this problem, and agrees more closely with human ratings than a single judge \citep{verga2024poll}.
	AlienBench therefore scores models with a panel of judges.
}%
%
%
\section{AlienBench}%
\label{sec:method}%
%
\paragraph{Task.}%
\label{sec:task}%
\final{%
	AlienBench adapts the structured imagination task of \citet{ward1994}.
	In this task, participants imagine and describe a creature that lives on a planet other than Earth.
	Each description is labeled on a set of biological features, such as symmetry and locomotion.
	A feature counts as a departure if it differs from the typical feature of Earth animals, such as bilateral symmetry or legs for locomotion.
	The number of departures measures how far a creature differs from Earth animals.
	\citet{ward1994} found that most creatures imagined by college students and by professional science-fiction writers share the typical
	features of Earth animals.
	This result is known as the structured imagination effect.
}%

\final{%
	AlienBench applies Ward's task 
	to language models.
	The model is prompted to generate a description of an alien creature in English.
	A panel of LLM judges then extracts \abNDimsWord{} biological features from the description, and labels the features as Earth-typical (0) or as a departure (1).
	The number of departures averaged over all the models generations is the Ward Departure Score, the primary metric of AlienBench.
	Two example descriptions of alien creatures are given in Appendix \ref{app:description_examples}.
}%

\begin{table}[htb]
	\caption{\ok{The \abNConditionsWord{} prompt conditions used in AlienBench. The prompt text is shown verbatim.}}
	\label{tab:prompts}
	\label{tab:conditions}
	\centering
	\scriptsize
	\begin{tabularx}{\linewidth}{@{}l>{\raggedright\arraybackslash}p{1.45cm}>{\raggedright\arraybackslash}X@{}}
  \toprule
  Acronym & Condition & Prompt text \\
  \midrule
  Base & Baseline & Imagine a creature that lives on an alien planet. Describe it in detail. \\
  Dep-P & Departure-primed & Imagine a creature on an alien planet that is as different from any Earth creature as possible. Describe it in detail. \\
  NoLight & Constrained: no light & Imagine a creature that lives on a planet with no light at all. Describe it in detail. \\
  HighG & Constrained: high gravity & Imagine a creature that lives on a planet with 10x Earth's gravity. Describe it in detail. \\
  NH$_3$ & Constrained: ammonia oceans & Imagine a creature that lives on a planet with oceans of liquid ammonia instead of water. Describe it in detail. \\
  Detail & Detailed description & Imagine a creature on an alien planet. Describe it in full detail, including any aspects of its biology that would help someone understand what kind of organism it is. \\
  \bottomrule
\end{tabularx}
\end{table}

\begin{table}[t]
	\caption{Biological features of the WDS.
	}
	\label{tab:ward}
	\centering
	\scriptsize
	\begin{tabularx}{\linewidth}{@{}l>{\raggedright\arraybackslash}p{2.0cm}>{\raggedright\arraybackslash}X@{}}
\toprule
Feature & Earth-typical default & Example departures \\
\midrule
Symmetry & Bilateral & Radial, asymmetric, fractal \\
Sensory organs & Eyes, ears, nose on head & Distributed sensing, EM reception \\
Locomotion & Limbs (2 or 4) & Sessile, rolling, jet propulsion \\
Body plan & Head, torso, limbs & No head, amorphous, colonial \\
Skin/covering & Skin, fur, scales, feathers & Crystalline, gaseous, energy-based \\
Reproduction & Sexual, binary sexes & Budding, spores, multiple sexes \\
Metabolism & Heterotrophic, aerobic & Chemosynthesis, radiation feeding \\
Communication & Acoustic or visual signals & Chemical, electromagnetic signaling \\
Habitat & Land or water surface & Subsurface, atmospheric, vacuum \\
Cognition & Centralized nervous system & Distributed, hive-level, non-neural \\
\bottomrule
\end{tabularx}
\end{table}

\paragraph{Prompt conditions.}%
\label{sec:prompts}%
\final{%
	Each language model is evaluated under \abNConditionsWord{} prompt conditions (see \autoref{tab:prompts}).
	The \textit{baseline} condition is a minimal open-ended instruction.
	%
	The \abNConstrainedWord{} \textit{constrained-environment} conditions specify a physically demanding planetary setting: no light (NoLight), \abGravityMultiplier{}$\times$ Earth gravity (HighG), or liquid ammonia oceans (NH\textsubscript{3}). 
	%
	The \textit{departure-primed} condition directlSectiony instructs the model to differ from Earth organisms as much as possible.
	%
	The \textit{detailed-description} condition retains the baseline framing but asks for a longer, more complete description without explicitly naming any biological aspect.
}

\final{%
	Each condition is run \abNPerCell{} times at different seed values (\abNPerModel{} generations per model). 
	A model's overall score is the mean WDS over all of its generations, after averaging over the \abNJudgesWord{} judges. 
	The paraphrase analysis in Appendix~\ref{sec:ablation_prompt} shows that scores are sensitive to the wording in the prompt.
	For this reason, 
	the benchmark is by design the average across all \abNConditionsWord{} conditions,
	as a deliberate design choice to measure departure across a range of task framings (cf. \autoref{fig:score_spread}).%
}%

\paragraph{Ward Departure Score (WDS).}%
\label{sec:ward}%
\final{%
The WDS counts how many of \abNDimsWord{} biological features depart from Earth-typical defaults: body symmetry, sensory organs, locomotion, body plan, skin or body covering, reproduction, metabolism, communication, habitat, and cognitive architecture (\autoref{tab:ward}).
	For each feature, a judge extracts the feature and scores it on a binary scale of 1 (departure) and 0 (Earth-typical default).
	A response scores \abScoreDeparture{} on a feature if it describes a different mechanism from that default, not a variation on it: `\abExampleLegsWord{} legs' scores \abScoreEarth{} on locomotion because legs remain the locomotion mechanism; `rolling' or `jet propulsion' scores~\abScoreDeparture{}.
	The extraction prompt provides boundary criteria that distinguish variation from departure.
	An Earth-typical default for each feature is given to the judge as an example (e.g., bilateral symmetry, locomotion via limbs, centralized nervous system).
	If a feature is missing in the description, it scores~0.
	This rule is part of the benchmark design and follows the present-or-absent coding of \citet{ward1994}.
	The full prompt is in \autoref{sec:extraction_prompt}.
}%

\final{%
	The WDS is the sum of the departure labels over the \abNDimsWord{} features (range 0--\abWardMax{}).
	Low scores on the benchmark indicate that a model's responses do not diverge from Earth-typical defaults.
High scores indicate that a model's responses depart further from defaults, but they do not indicate that the descriptions are more coherent or better written.
	Response length is a decisive factor in this benchmark.
	A long response 
	has a higher probability to depart from Earth-typical defaults, following \citet{ward1994}.
	The score, therefore, reflects both the level of detail of a description and the structural novelty of the features it specifies.
}%


\final{%
	The departure label of each feature is binary
	for three reasons.
	First, it follows the present-or-absent coding of \citet{ward1994}.
	Second, a binary label is easy to interpret. A continuous label would require a means to measure between qualitatively different mechanisms (e.g., how far `rolling' lies from `walking on legs').
	A binary label reduces each judgment to one decision at an explicit, well-defined boundary (see \autoref{tab:boundary} in Appendix \ref{sec:extraction_prompt}).
	These boundaries were found to be tractable for LLM judges in pilot testing compared to a continuous score.
	Third, the overall score of a model is an aggregate over its generations, and is thus continuous: the panel mean and the mean of per-feature departure both lie on a continuous scale from 0 and 1.
}%

\paragraph{Models.}%
\label{sec:models}%
\final{%
We evaluate \abNModelsWord{} language models, including \abNClosedWord{} proprietary frontier models and \abNOpenWord{} open-weight models 
(see Appendix~\ref{app:models}).
The models generate the alien creature descriptions with temperature \abTemperature{} and at most \abMaxTokens{} output tokens (\abMaxTokensLlamaMaverick{} for \abMaxTokensModelLlamaMaverick{}),
with $n = \abNPerCell{}$ independent responses per model per condition (\abNPerModel{} per model).
\todo{All responses stayed below the output token cap (Max: \abMaxCapSharePct{}\%).}%
}%

\final{%
	The judge pool is a fixed panel of \abNJudgesWord{} LLM judges:
	\abJudgePanel{}.
	Judge extraction uses temperature \abJudgeTemperature{}\rev{ and at most \abJudgeMaxTokens{} output tokens, doubled up to \abJudgeMaxTokensCeiling{} if a response is cut off}.
	\rev{A judge output without a valid label for every feature is requested again, up to \abMaxParseAttemptsWord{} attempts in total.
	No judge call reached its token limit, and no judge output needed a second attempt.}
	Each response is scored independently by every judge, and a mean score is calculated from the three ratings.
}%

\paragraph{Human baseline.}%
\label{sec:human}%
\ok{%
	To establish a human baseline for the WDS, we recruited \abHumanN{} participants through \abHumanPlatform{}\footnote{\url{https://www.prolific.com}}
	(
	minimum \abHumanApproval{}\% approval rate, at least \abHumanPriorTasks{} prior accepted tasks, and English as first language).
	\todo{Three participants were excluded (see Limitations) and filled with other participants.}
	Tasks were released in small batches at different times of the day and different days of the week.
}%

\ok{%
	Participants were \abHumanFemale{}\% female and \abHumanMale{}\% male, with a median age of \abHumanAgeMedian{} years (range \abHumanAgeMin{}--\abHumanAgeMax{}).
	\abHumanCountryOneNCap{} participants resided in \abHumanCountryOne{}, \abHumanCountryTwoN{} in the \abHumanCountryTwo{}, and \abHumanCountryRestN{} in \abHumanCountryRestWord{} other countries.
	\abHumanBioGradNCap{} participants reported a graduate or professional background in biology or life sciences, \abHumanBioUndergrad{} familiarity from undergraduate and \abHumanBioSchool{} from school-level courses, and \abHumanBioNone{} had no background.
	\abHumanPriorYesCap{} participants reported that they had previously taken part in a study that asked them to describe or draw an imaginary alien creature.
}%

\final{%
	\todo{Based on a small pilot, the study was advertised at \abHumanEstMinutes{}~min with a reward of \abHumanReward{} (\abHumanAdvertisedRate{}/h), and participants received \abHumanPayPerParticipant{} on average.
	At a median completion time of \abHumanMinutes{}~min \abHumanSeconds{}~sec, the pay to participants corresponds to \abHumanPay{} per hour.}
	Each participant completed the baseline prompt once and was instructed not to consult AI tools or use web search.
	Appendix~\ref{app:human_study} reproduces the full study text.
	Responses were scored with the same extraction and scoring pipeline as the model outputs.
	The results of the human baseline are reported 
	in Section~\ref{sec:human_comparison}.%
}%

\paragraph{Statistical analysis.}%
\label{sec:analysis}%

\final{%
	The overall model scores 
	are means over all generations from the \abNConditionsWord{} conditions, after averaging
	the judges for each generation.
	Scores are reported with bootstrap percentile \abCILevel{}\% CIs ($B=\abBootstrapB{}$).
	The comparison of open-weight and proprietary models uses the model as the unit of analysis.
	An exact permutation test compares the mean overall scores of the two groups across all \abOpenClosedNSplits{} assignments of the models to groups of their original sizes.
	The CI of this difference comes from a two-level bootstrap that resamples models within each group and generations within each model.
	We assess inter-rater reliability among the panel of \abNJudgesWord{} LLM judges with Krippendorff's alpha \citep{krippendorff2011}. 
	Agreement is interpreted according to the 
	bands by \citet{landis1977}, with good inter-judge agreement being the lowest acceptable value ($\alpha \geq \abAlphaGood{}$).
	Appendix~\ref{app:implementation} lists the software that implements these analyses.%
}%

\ok{%
	Differences in the WDS across prompt conditions are tested with a Kruskal-Wallis test.
	The test uses one score per generation, obtained by averaging the judges, and pools generations across models.
	Post-hoc comparisons use two-sided Mann-Whitney $U$ tests between all pairs of conditions with Bonferroni correction ($\alpha' = \abBonferroniAlpha{}$).
	We report the comparisons between the baseline condition and each other condition.
	Effect sizes are rank-biserial correlations with bootstrap percentile \abCILevel{}\% CIs ($B=\abBootstrapB{}$).
	A positive correlation indicates a higher score than under the baseline condition.
	A Monte Carlo sensitivity analysis with simulated normal location shifts estimates the minimum detectable effect at \abPowerTarget{}\% power, given the realized sample size and $\alpha'$.%
}%

\ok{%
	Longer response texts may describe more features and thus contain more departures.
	AlienBench therefore additionally reports the WDS 
	per \abTokensPerRateUnit{} completion tokens (\autoref{tab:ward_scores}). 
}%


%
%
%
\section{Experiments}%
\label{sec:results}%
%
\subsection{Model Comparison}%
\label{sec:comparison}%
%
\begin{table*}[thbp]
	\centering
	\caption{%
		\ok{%
			Ward Departure Scores (0--\abWardMax{}) by model and prompt condition.
			The first column is the model's overall WDS, averaged over the scores across the \abNConditionsWord{} conditions.
			This column reports the per-model mean with a bootstrap percentile \abCILevel{}\% CI  ($B=\abBootstrapB{}$).
			The second column reports the WDS per \abTokensPerRateUnit{} completion tokens. 
			The remaining columns report the mean (SD) across generations after averaging the judges' scores.
		}%
	}
	\small
	\label{tab:ward_scores}
	\resizebox{\textwidth}{!}{%
		\begin{tabular}{lcccccccc}
  \toprule
  Model & Overall WDS & WDS / 100 token & Base & Dep-P & NoLight & HighG & NH$_3$ & Detail \\
  \midrule
    Claude Opus 5.5 & 5.14\,[4.78, 5.51] & 0.16\,[0.15, 0.17] & 4.26 (1.50) & 9.82 (0.31) & 3.68 (1.47) & 2.69 (0.61) & 5.57 (1.14) & 4.84 (1.53) \\
    GPT-6 Astra & 4.36\,[3.98, 4.77] & 0.43\,[0.40, 0.47] & 3.07 (1.04) & 9.61 (0.43) & 2.80 (0.66) & 2.31 (0.76) & 5.02 (0.83) & 3.36 (1.79) \\
    GPT-6 Luna & 3.28\,[2.87, 3.70] & 0.30\,[0.27, 0.33] & 2.16 (1.14) & 8.82 (0.61) & 2.51 (0.93) & 0.34 (0.51) & 3.67 (1.10) & 2.16 (1.51) \\
    Gemini 3.8 Flash & 4.72\,[4.39, 5.06] & 0.19\,[0.18, 0.21] & 3.20 (0.98) & 9.19 (0.50) & 2.37 (0.85) & \textbf{4.18 (0.74)} & 4.44 (0.67) & 4.92 (1.27) \\
    Grok 4.7 & 5.42\,[5.01, 5.84] & \textbf{0.60\,[0.55, 0.65]} & 4.56 (1.75) & 9.31 (0.52) & 4.68 (1.41) & 1.24 (0.87) & \textbf{5.94 (0.82)} & 6.79 (2.03) \\
  \midrule
    DeepSeek V4.1 Flash & 4.97\,[4.54, 5.41] & 0.09\,[0.09, 0.10] & 3.18 (1.98) & \textbf{9.87 (0.30)} & 3.79 (1.48) & 2.16 (1.07) & 5.02 (1.22) & 5.82 (2.48) \\
    Llama 4 Maverick & 4.84\,[4.45, 5.22] & 0.60\,[0.54, 0.65] & 3.99 (1.99) & 8.68 (0.70) & 5.31 (1.20) & 1.97 (1.38) & 5.52 (1.48) & 3.54 (2.21) \\
    Mistral Large 4 & 5.41\,[5.03, 5.81] & 0.15\,[0.14, 0.16] & 5.16 (2.20) & 9.42 (0.55) & 4.26 (1.23) & 2.38 (1.41) & 5.50 (1.42) & 5.77 (2.38) \\
    Qwen3.8 & 5.23\,[4.76, 5.69] & 0.16\,[0.13, 0.19] & 4.54 (2.28) & 9.81 (0.29) & 3.94 (1.49) & 1.29 (0.51) & 5.47 (1.66) & 6.34 (3.13) \\
    Ministral 3 & \textbf{6.55\,[6.19, 6.89]} & 0.36\,[0.33, 0.38] & \textbf{6.88 (2.04)} & 8.74 (0.67) & \textbf{7.11 (1.63)} & 3.40 (2.45) & 5.89 (1.70) & \textbf{7.27 (2.37)} \\
  \bottomrule
\end{tabular}
	}%
\end{table*}

\ok{%
	Across the \abNModelsWord{} models, the overall WDS ranges from \abWorstMeanCI{} for \abWorstModelSized{} to \abBestMeanCI{} for \abBestModelSized{} on the 0--\abWardMax{} scale (\autoref{tab:ward_scores}).
	The highest score under each condition comes from \abNCondLeadersWord{} different models, \abNCondLeadersOpenWord{} of them open-weight models. 
	%
}%
\ok{%
	The open-weight models score higher than proprietary models, on average.
	The mean overall score is \abOpenMean{} for the 
	open-weight models and \abClosedMean{} for the 
	proprietary models.
	\rev{The difference of \abOpenClosedDiff{} Ward points is not statistically significant (permutation test, \abOpenClosedPermP{}, \abCILevel{}\% CI \abOpenClosedDiffCI{}).}
	\rev{This comparison is descriptive.
	With \abNOpenWord{} models per group, the test has limited power to detect a difference (see Limitations).}
	The open-weight model \abOpenAboveClosedModelsSized{} scores higher than the best proprietary model, \abBestClosedModelSized{}. 
	\abWorstClosedModelSized{} (\abWorstClosedMean{}) scores lowest among the evaluated models.
}%

\ok{%
\abDensestModelSized{} has the most departures per \abTokensPerRateUnit{} completion tokens (\abDensestRate{}).
Its descriptions are shorter than those of \abBestModelShort{}, the model with the highest overall score (mean of \abDensestTokens{} vs.\ 
\abBestModelTokens{} completion tokens).
\abSparsestModelShort{} has the fewest departures per \abTokensPerRateUnit{} completion tokens (\abSparsestRate{}), which indicates that its
descriptions are long relative to the number of departures they contain.
}%

\begin{figure}[htbp]
\centering
\includegraphics[width=.95\linewidth]{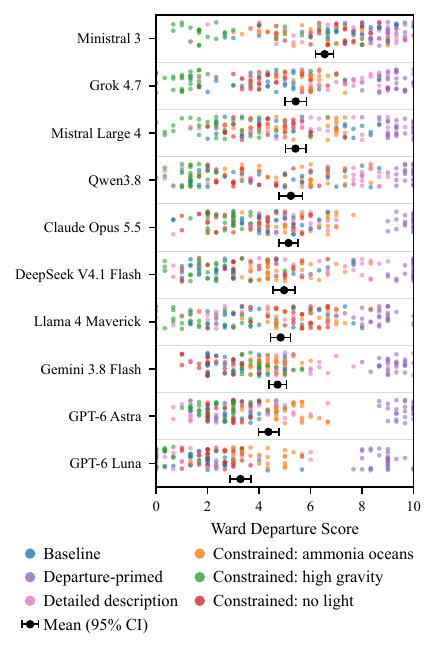}
\caption{\ok{Distribution of Ward Departure Scores (0--\abWardMax{}).
		Each point is one generation, averaged across the \abNJudgesWord{} judges, under different prompt conditions.
			The black dot 
			marks the mean over \abNPerModel{} generations, and whiskers mark the bootstrap \abCILevel{}\% CI ($B=\abBootstrapB{}$).
	}}
	\label{fig:score_spread}
\end{figure}

\ok{%
	\autoref{fig:score_spread} shows the spread of scores across the conditions.
	All models score high under the departure-primed condition (\abCondMinDeparturePrimed{} or higher).
	Even low scoring models can perform well under the departure-primed condition.
	%
	\abBestModelSized{} has the highest overall score, \abLeadMargin{} Ward points above \abSecondModels{}.
	Two examples of alien creature descriptions are given in \autoref{app:description_examples}.%
}%

\ok{%
	Scores differ across the \abNConditionsWord{} prompt conditions (Kruskal-Wallis test, $H = \abKruskalH{}$, \abKruskalP{}).
	Pairwise Mann-Whitney $U$ with Bonferroni correction shows lower scores under high gravity than under the baseline ($r =
	\abRConstrainedHighGravity{}$, \abCILevel{}\% CI \abRCIConstrainedHighGravity{}, \abPConstrainedHighGravity{}).
	Scores are higher under the departure-primed condition ($r = \abRDeparturePrimed{}$, \abCILevel{}\% CI \abRCIDeparturePrimed{},
	\abPDeparturePrimed{}), under ammonia oceans ($r = \abRConstrainedAmmonia{}$, \abCILevel{}\% CI \abRCIConstrainedAmmonia{},
	\abPConstrainedAmmonia{}), and under the detailed-description condition ($r = \abRDetailedDescription{}$, \abCILevel{}\% CI
	\abRCIDetailedDescription{}, \abPDetailedDescription{}).
	There is no detectable difference for the no-light condition ($r = \abRConstrainedNoLight{}$, \abCILevel{}\% CI \abRCIConstrainedNoLight{},
	\abPConstrainedNoLight{} after correction).
	The minimum detectable effect is $|r| = \abMinDetectableR{}$.
}%

\begin{figure*}[thbp]
	\centering
	\includegraphics[width=.9\linewidth]{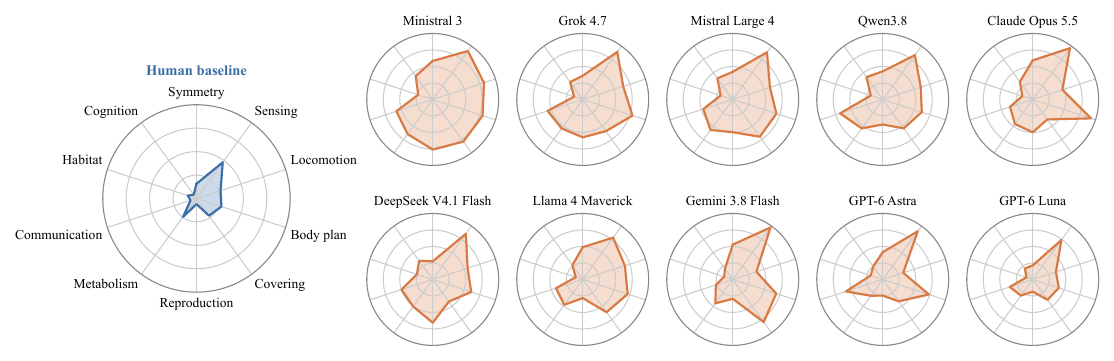}
	\caption{\ok{Departure rate of each of the \abNDimsWord{} features, per model, in descending order of Ward Departure Score. 
			The distance from the center on each axis is the departure rate of that feature.
	}}
	\label{fig:ward_radar}
\end{figure*}

\subsection{Comparison to Human Baseline}%
\label{sec:human-comparison}%
\label{sec:human_comparison}%
\final{%
	\autoref{fig:ward_radar} compares the models with the human baseline.
	The exact values are listed in \autoref{tab:ward_dimensions} in Appendix~\ref{app:ward_dimensions}.
	In the baseline condition, the alien creature descriptions collected from crowdsourced human participants score low on feature departure, lower than the worst performing model.
}%

\subsection{Reliability of the LLM Judge Panel}%
\label{sec:judge-reliability}%
%
\final{%
	All \abNExtractionCalls{} judge calls returned valid JSON.
	Inter-rater reliability of the LLM judge panel is reported in \autoref{tab:reliability_full} in Appendix~\ref{app:reliability}.
	The Ward Departure Score reaches Krippendorff's $\alpha = \abAlphaWard{}$, which indicates \MakeLowercase{\abAlphaWardBand{}} agreement among the \abNJudgesWord{} judges \citep{landis1977}.
	\abNExcellentDimsCap{} features fall in the Excellent band, \abNGoodDimsWord{} in the Good band, and \abNModerateDimsWord{} in the Moderate band.
	For the departure labels, \abNClearDimsWord{} of the \abNDimsWord{} features clear the $\alpha \geq \abAlphaGood{}$ good-agreement threshold, and only \abBelowGoodDims{} falls below it ($\alpha = \abBelowGoodMin{}$).
}%
%
%
%
\section{Ablations and Sensitivity Analyses}%
\label{sec:ablations}%
%
\final{We report three ablations of the benchmark design.
	Appendix~\ref{app:ablations} provides the detailed procedures, statistics, and result tables.%
}%

\paragraph{Prompt paraphrase sensitivity.}
\ok{%
	We test whether the results 
	depend on the wording of the prompt by
	generating \abParaN{} responses per model under \abNParaphrasesWord{} hand-written rewordings (\autoref{tab:paraphrases} in Appendix \ref{sec:paraphrase_sensitivity}).
	Model rankings are partly preserved across wordings, and absolute scores vary under rewordings.
	The median Spearman correlation between the rankings from two wordings is $\rho = \abParaRhoMedian{}$.
	\todo{Three quarters (\abParaNAboveNull{} of \abParaNPairs{}) of the ordered wording pairs exceed the upper bound of a permutation null, but only \abParaNRhoBarWord{}
	reach the pre-specified bar of $\rho \geq \abParaRhoBar{}$.
	Across wording pairs, model scores differ by a median of \abParaMadMedian{} Ward points.}
	Therefore, benchmark runs should use the exact prompt wording, and scores or rankings from different wordings should not be compared directly.
	Appendix~\ref{sec:ablation_prompt} provides the detailed results.%
}%
\paragraph{Feature sensitivity.}
\final{%
	We test whether the results depend on the set of \abNDimsWord{} features.
	We remove one feature at a time and recompute the score from the remaining \abNDimsMinusOneWord{} features.
	The resulting model rankings correlate with the
	\abNDimsWord{}-feature ranking 
	at $\rho \in [\abDimLooRhoMin{},
	\abDimLooRhoMax{}]$.
	The model ranking therefore does not depend on any single feature.
}%
\ok{%
	We also rank the models by each feature alone.
	Only `\abDimBestSingleDim{}' reaches the pre-specified bar of $\rho \geq \abDimRhoBar{}$ for the correlation with the same ranking ($\rho =
	\abDimBestSingleRho{}$).
	\todo{%
	This feature was selected after the fact on the same \abNModelsWord{} models, so the correlation may not hold for other models.}
	The correlation also compares only the order of the models, and a single feature does not cover the breadth of departure that the benchmark measures.
	Absolute scores depend on the number of features because the score sums over them.
	Even the best-matching subset differs from the \abNDimsWord{}-feature score by less than \abParaMadBar{} Ward points on average only if it contains at least \abDimMadBelowOneK{} features.
	Appendix~\ref{sec:ablation_dimensions} provides the detailed results.%
}%

\paragraph{Judge panel composition.}
\final{%
	We test whether the results depend on the composition of the judge panel.
	The results change little when judges are removed.
	The ranking from each single judge correlates with the ranking from the remaining judges at $\rho \in [\abJudgeSingleRhoMin{},
	\abJudgeSingleRhoMax{}]$, and 
	all \abNJudgesWord{} judges reach the bar of $\rho \geq
	\abJudgeRhoBar{}$.
	Panels without one judge have a Krippendorff's $\alpha$ between \abJudgeLooAlphaMin{} and \abJudgeLooAlphaMax{}.
	We also test for self-preference bias, which would predict a lower model rank after removal of the judge from the same model family.
	\abSfNPairsCap{} models have a judge from the same model family on the panel.
	All of these models keep the same rank after removal of that judge.
	This result provides no 
	evidence of self-preference bias in model ranks, but it does not exclude smaller differences in scores.
	Appendix~\ref{sec:ablation_judges} provides the detailed results.
}%

\section{Discussion}%
\label{sec:discussion}%

\final{%
	AlienBench is a benchmark for measuring structural novelty in language model generations.
	The Ward Departure Score quantifies the departure in alien creature descriptions from earth-typical defaults.
	This benchmark differs from prior LLM creativity benchmarks in three respects.
	First, it scores outputs with a 
	rubric from a validated task from Cognitive Psychology. 
	Second, it evaluates models under constrained generation, with prompts that specify the physical environment of the planet, and averages scores across \abNConditionsWord{} different conditions.
	Third, the benchmark considers output length as one determinant of model creativity: a model that outputs one sentence will score lower than a model that outputs several pages of text.
}%


\final{%
	Models differ in their ability to generate structural novelty in their outputs.
}%
	\rev{Among the evaluated models, the highest overall score comes from an open-weight model, but the difference between the mean scores of open-weight and proprietary models is not statistically significant.}
	However, no single model has the highest departure rate on every feature.
	The models differ in which features they depart on, and 
	the radial profiles in \autoref{fig:ward_radar} differ in shape, not only in scale.
	The Overall WDS reflects these differences in one number.%

\ok{%
	In the baseline condition, models depart from Earth-typical defaults more than humans on every feature, on average.
	One possible explanation is that the training data includes descriptions of alien creatures, such as science-fiction prose.
	The WDS cannot test this explanation because it measures whether a description departs from Earth-typical defaults, not how the description was produced.
	We therefore report the human baseline as a descriptive reference and not as a comparison of the internal creative processes in humans and models.%
}%

\final{%
	The score difference between language models and human participants should be interpreted in light of the conflation between originality and elaboration in the benchmark.%
Text production requires more time and effort from human participants than from language models, and this difference may contribute to the observed score gap.
The crowdsourcing model may also contribute to this result.
Crowdsourced participants are extrinsically motivated to complete tasks as quickly as possible to maximize their monetary compensation. This means participants write less text, and 
score low on feature departure.
We do not consider this comparison a final human baseline, and future work could investigate the difference between human and model creativity in more depth.
}%


\subsection*{Is the structured imagination effect present in language models?}%
\label{sec:discussion_effect}%
%
\final{%
	\citet{ward1994} found that people often reuse features of Earth animals when they imagine creatures from another planet.
	AlienBench's baseline condition matches the task of \citet{ward1994}, and the 
	WDS thus can be used to 
	test the existence of the structured imagination effect in language models.
}%

\final{%
	Under the baseline condition, most models show the structured imagination effect, but they perform better on this benchmark than the human participant sample.
	The human participants score only \abHumanMean{} on average (\abCILevel{}\% CI \abHumanMeanCI{}), which means the features mentioned by participants do not, on average, depart from Earth-typical defaults (or the features are absent in the alien creature descriptions).
	The models 
	score between \abCondMinBaseline{} to \abCondMaxBaseline{} on the baseline prompt.
	This means the models' generations show less of the structured imagination effect than the human-written texts.%
}%
\section{Conclusion}%
\label{sec:conclusion}%
%
\final{%
AlienBench measures structural novelty in language models' descriptions of alien creatures, scored by a panel of LLM judges with high
agreement.
Model scores vary widely, and no single model scores highest under every prompt condition.
Under the baseline condition, all models depart from Earth-typical defaults more often than the human participant sample.
}%

	
\ok{%
	The ablations show that model rankings depend little on any single feature or judge, but scores vary with the prompt wording and the
	number of features.
	AlienBench is therefore intended for relative comparisons across models under its fixed configuration.
	Future models can be evaluated with the same prompts, features, and judge panel as the \abNModelsWord{} models reported here.
	AlienBench's protocol of theory-grounded features, binary departure labels, and explicit boundary criteria may also apply to other generative tasks with known defaults, such as design fixation tasks \citep{JANSSON19913,Smith1993}.
}%

\clearpage

\section*{Limitations}%
\label{sec:limitations}%
%
\final{%
	\paragraph{Scope.}
	AlienBench measures structural novelty (i.e., departure in kind from Earth-typical defaults) in a single domain. It does not measure divergent creativity in the Guilford--Torrance tradition~\citep{runco2012standarddef}, nor creative quality as found in open-ended writing, design, or problem solving. Ward Departure Scores should not be used as a proxy for general creative ability.
	Section~\ref{sec:ablations} shows that scores vary with the prompt wording and the number of features.
	A model's Ward Departure Score is a mean over these \abNConditionsWord{} prompts.
	AlienBench is intended for relative comparisons across models under this fixed configuration of prompts, features, and Earth-typical defaults.
}%

\final{%
	\paragraph{Exposure in training data.}
	One limitation of this benchmark is that language models have likely encountered Ward's task \cite{ward1994} and its exemplars during training, together with decades of science-fiction text describing alien creatures. 
	Therefore, a model's performance on AlienBench may reflect memorized tropes instead of structural novelty.
	Binary labels compare the alien creatures with the Earth-typical defaults, but they do not distinguish a memorized creature description from a newly imagined one.
	This is a general problem of text-based creativity benchmarks where stochastic outputs of language models resemble creativity on a surface level, but model creativity is indistinguishable from mere reproduction of learned patterns acquired during pretraining.%
}%

\final{%
	\paragraph{Language and culture.}
	All prompts and responses are in English.
	The Earth-typical defaults reflect a biological taxonomy common in English-language biology textbooks and science fiction.
	The benchmark may therefore measure departure from a culturally specific reference, and it does not cover other languages.%
}%

\final{%
	\paragraph{Reproducibility.}
	Each generation call carries a fixed seed, but providers honor seeds only on a best-effort basis, and outputs can change with the serving infrastructure or the model snapshot.
	A rerun of the pipeline may therefore produce different scores, whereas analyses of the released generations and judge outputs are exactly reproducible.
	Scoring new models with the same panel requires the same judge snapshots, which providers may retire.%
}%

\rev{%
	\paragraph{Model sample.}
	The evaluated models were selected to cover the major model providers and a range of model sizes.
	They were not selected to test whether open-weight and proprietary models differ.
	The comparison of the two groups treats each model as one observation, with \abNOpenWord{} models per group.
	The exact permutation test on this sample has \abOpenClosedNSplits{} possible group assignments and can detect only large differences between the group means.
	A non-significant result in this comparison is therefore not evidence that the two groups perform equally.
	The released benchmark allows the comparison to be repeated with a larger set of models.%
}%

\final{%
	\paragraph{Human baseline.}
	Another limitation is the use of crowd workers who are known to use LLMs to complete tasks
	\citep{Evgenia,rilla2025recognisinganticipatingmitigatingllm,10.1145/3685527}.
	While we excluded \abHumanExcludedWord{} suspicious responses 
	 that included markdown formatting and unusual subheadings, more subtle uses could not be detected from text alone. The human baseline reported in Section~\ref{sec:human_comparison} should be interpreted with this limitation in mind.%
}%

\final{%
	\paragraph{LLM judges.}
	AlienBench uses a panel-of-judges design \cite{verga2024poll} in which each judge scores all models, including models from the same provider family as the judge.
	The self-preference bias of \citet{panickssery2024selfpreference}
	may therefore be present in one judge when rating same-family models. 
	The overall score is the mean over all judges, which
	limits this judge's share of each score to one third but does not fully remove the potential same-family bias \cite{verga2024poll}.
	The panel composition can also not remove agreement arising from shared pretraining corpora across all judges and models. Overlapping text corpora may produce agreement for reasons unrelated to the labeling rules. 
	We report on a leave-one-judge-out sensitivity analysis in Appendix~\ref{sec:ablation_judges}.
}%

\section*{Ethics Statement}%
\label{sec:ethics}%
%
\final{%
	\paragraph{Risks.}
	AlienBench scores fictional descriptions of alien creatures, and we see no direct path to misuse.
	One risk is optimization for the benchmark itself.
	The score is associated with response length, 
	so tuning a model toward a higher score may reward longer or more unusual text without better writing.
	A second risk is overgeneralization from the human comparison.
	This comparison concerns one prompt and one crowdsourced sample (N = 100), and it does not show that language models are more creative than people in general.
	The study used API inference only, within the providers' terms of use, and trained no models. 
}%

\final{%
	\paragraph{Ethics review.}
	Under the regulations of our institution and country, 
	this study is exempt from review by an Institutional Review Board.
}%

	\paragraph{Personal information and offensive content.}
	Personal information was stripped from the data before release.
Participants gave informed consent and were told their alien creature descriptions would be released publicly.	
	The released human responses contain no participant identifiers.
	We remove the \abHumanPlatform{} IDs before release, and the released data contain no demographic information.
	A script searched all human and model texts for e-mail addresses, URLs, phone numbers, user handles, and \abHumanPlatform{} IDs and found none.
	We also read all human responses.
	None of them names or describes an identifiable person.
	\abHumanNOffensiveCap{} response contains an ableist term, and some responses describe graphic violence within the fictional scenario.
	We release these responses unchanged because they are part of the human baseline\rev{, which is intended only as a research reference for model evaluation}.
	A keyword search for profanity and slurs found no offensive terms in the model generations.

\bibliography{paper}

\clearpage

\appendix


\final{%
	\section{Data Card}
	\label{app:datacard}
	
	This appendix documents the released AlienBench data in the form of a data card \citep{pushkarna2022datacards}.
	The data card follows the themes of the Data Card template.
	Themes that do not apply to an evaluation benchmark of generated text, such as retention, wipeout, and transformations, are omitted.%
	\paragraph{Summary.}
	AlienBench contains free-form English descriptions of alien creatures written by language models and by human participants, the feature labels that an LLM judge panel assigns to each description, and the resulting scores.
	The data support the comparison of language models on structural novelty in descriptions of alien creatures.

	\paragraph{Authorship.}
	The authors are withheld for anonymous review.

	\paragraph{Dataset overview.}
	\begin{itemize}[nosep, leftmargin=*]
		\item \textbf{Data subjects.}
		The texts describe fictional creatures that live on planets other than Earth.
		The human responses are written by crowd workers.
		
		\item \textbf{Content description.}
		The release contains four record types: model generations, human responses, judge outputs, and scores.
		
		\item \textbf{Descriptive statistics.}
		The main study contains \abNGenerations{} generations from \abNModelsWord{} language models (Appendix~\ref{app:models}).
		Each model contributes \abNPerCell{} generations under each of the \abNConditionsWord{} prompt conditions (\autoref{tab:prompts}).
		The prompt paraphrase analysis adds \abParaN{} generations per model under each of \abNParaphrasesWord{} paraphrases (\autoref{tab:paraphrases}).
		The human baseline contains \abHumanN{} responses to the baseline prompt.
		Each of the \abNJudgesWord{} judges labels every generation, which yields \abNExtractionCalls{} extraction records for the main study.
		
		\item \textbf{Sensitivity of data.}
		The human responses contain no participant identifiers and no demographic data.
		The Ethics Statement describes the screening for personal information and offensive content.
	\end{itemize}

	\paragraph{Example of data points.}
	\begin{itemize}[nosep, leftmargin=*]
		\item \textbf{Primary data modality.}
		Text.
		
		\item \textbf{Data fields.}
		A generation record contains the requested model, the resolved model snapshot, the prompt, the response, the sampling temperature and seed, the token counts, and a timestamp.
		A human record contains the response and the answers to the two final questions of the study form (Appendix~\ref{app:human_study}).
		A judge record contains, for each of the \abNDimsWord{} features, a short description of the feature, a binary departure label, and a one-sentence rationale (Appendix~\ref{sec:extraction_prompt}).
		A score record contains the Ward Departure Score and the departure label of each feature for one generation and one judge.
		
		\item \textbf{Typical data points.}
		Appendix~\ref{app:description_examples} reproduces two model generations.
	\end{itemize}

	\paragraph{Motivations and intentions.}
	\begin{itemize}[nosep, leftmargin=*]
		\item \textbf{Purpose.}
		The data support relative comparisons of language models on structural novelty under the fixed configuration of prompts, features, and judges.
		
		\item \textbf{Unsuitable use cases.}
		The scores are not a measure of general creative ability (see Limitations).
		The human responses are intended only as a research reference for model evaluation.
	\end{itemize}

	\paragraph{Access.}
	The data are available on Hugging Face (Section~\ref{sec:introduction}).
	The data are released under CC-BY 4.0, which the providers' terms of use permit for model outputs.
	The code is MIT-licensed.

	\paragraph{Provenance.}
	\begin{itemize}[nosep, leftmargin=*]
		\item \textbf{Model generations.}
		The evaluated models were accessed through OpenRouter, each pinned to a single upstream provider (Appendix~\ref{app:models}).
		Each generation call carries a fixed seed.
		
		\item \textbf{Human responses.}
		Participants were recruited through \abHumanPlatform{} with a minimum approval rate of \abHumanApproval{}\%, at least \abHumanPriorTasks{} prior accepted tasks, and English as first language.
		Each participant answered the baseline prompt once.
		
		\item \textbf{Data exclusion.}
		The human baseline excludes \abHumanExcludedWord{} responses that contained markdown formatting and unusual subheadings that suggest the use of a language model.
	\end{itemize}

	\paragraph{Human and other sensitive attributes.}
	The release contains no sensitive human attributes.
	Participants gave informed consent and were told that their descriptions would be released publicly.

	\paragraph{Annotations and labeling.}
	The labels are produced by a panel of \abNJudgesWord{} LLM judges (\abJudgePanel{}) with the extraction prompt in Appendix~\ref{sec:extraction_prompt}.
	The release contains no human annotations.

	\paragraph{Known applications and benchmarks.}
	AlienBench is an evaluation benchmark.
	The data have no training, validation, or test splits.
	\autoref{tab:ward_scores} reports the scores of the evaluated models.
}%


\section{Models Evaluated}
\label{models:evaluated}
\label{app:models}
\ok{%
	The evaluated models are listed below. 
	The resolved version is the dated provider snapshot returned by OpenRouter\rev{\footnote{\url{https://openrouter.ai}}} with the
	response.
	The parameter counts are those stated by the providers.
	For mixture-of-experts models, the list gives the total count and the count of parameters active per token.
	The providers of the proprietary models do not disclose parameter counts.
}%


\begingroup
\small
\setlist{nosep}
\raggedright
\begin{itemize}
  \item Claude Opus 5.5
  \begin{itemize}
    \item Provider: Anthropic
    \item OpenRouter slug: \href{https://openrouter.ai/anthropic/claude-opus-5.5}{\nolinkurl{anthropic/claude-opus-5.5}}
    \item Resolved version: \nolinkurl{anthropic/claude-opus-5.5-20260921}
  \end{itemize}
  \item GPT-6 Astra
  \begin{itemize}
    \item Provider: OpenAI
    \item OpenRouter slug: \href{https://openrouter.ai/openai/gpt-6-astra}{\nolinkurl{openai/gpt-6-astra}}
    \item Resolved version: \nolinkurl{openai/gpt-6-astra-20260903}
  \end{itemize}
  \item GPT-6 Luna
  \begin{itemize}
    \item Provider: OpenAI
    \item OpenRouter slug: \href{https://openrouter.ai/openai/gpt-6-luna}{\nolinkurl{openai/gpt-6-luna}}
    \item Resolved version: \nolinkurl{openai/gpt-6-luna-20260922}
  \end{itemize}
  \item Gemini 3.8 Flash
  \begin{itemize}
    \item Provider: Google
    \item OpenRouter slug: \href{https://openrouter.ai/google/gemini-3.8-flash}{\nolinkurl{google/gemini-3.8-flash}}
    \item Resolved version: \nolinkurl{google/gemini-3.8-flash-20260902}
  \end{itemize}
  \item Grok 4.7
  \begin{itemize}
    \item Provider: xAI
    \item OpenRouter slug: \href{https://openrouter.ai/x-ai/grok-4.7}{\nolinkurl{x-ai/grok-4.7}}
    \item Resolved version: \nolinkurl{x-ai/grok-4.7-20260916}
  \end{itemize}
  \item DeepSeek V4.1 Flash 1017
  \begin{itemize}
    \item Provider: DeepSeek
    \item OpenRouter slug: \href{https://openrouter.ai/deepseek/deepseek-v4.1-flash}{\nolinkurl{deepseek/deepseek-v4.1-flash}}
    \item Resolved version: \nolinkurl{deepseek/deepseek-v4.1-flash-20260910}
    \item Parameters: 552B total, 8B active on input and 16B on output
  \end{itemize}
  \item Llama 4 Maverick
  \begin{itemize}
    \item Provider: Meta
    \item OpenRouter slug: \href{https://openrouter.ai/meta-llama/llama-4-maverick}{\nolinkurl{meta-llama/llama-4-maverick}}
    \item Resolved version: \nolinkurl{meta-llama/llama-4-maverick-17b-128e-instruct}
    \item Parameters: 400B total, 17B active
  \end{itemize}
  \item Mistral Large 4
  \begin{itemize}
    \item Provider: Mistral AI
    \item OpenRouter slug: \href{https://openrouter.ai/mistralai/mistral-large-4-0}{\nolinkurl{mistralai/mistral-large-4-0}}
    \item Resolved version: \nolinkurl{mistralai/mistral-large-4-0-20261006}
    \item Parameters: 1T total, 52B active
  \end{itemize}
  \item Qwen3.8 27B fp8
  \begin{itemize}
    \item Provider: Alibaba
    \item OpenRouter slug: \href{https://openrouter.ai/qwen/qwen3.8-27b}{\nolinkurl{qwen/qwen3.8-27b}}
    \item Resolved version: \nolinkurl{qwen/qwen3.8-27b-20260814}
    \item Parameters: 27B (dense)
  \end{itemize}
  \item Ministral 3 14B 2512
  \begin{itemize}
    \item Provider: Mistral AI
    \item OpenRouter slug: \href{https://openrouter.ai/mistralai/ministral-14b-2512}{\nolinkurl{mistralai/ministral-14b-2512}}
    \item Resolved version: \nolinkurl{mistralai/ministral-14b-2512}
    \item Parameters: 14B
  \end{itemize}
\end{itemize}
\par
\endgroup


\section{Extraction Prompt}
\label{sec:extraction_prompt}

\ok{%
	This section provides the full prompt used by each judge to label the \abNDimsWord{} features of a creature description.
	The prompt consists of two parts: a system and a user prompt.
}%

\paragraph{System prompt.}%
\final{%
	{\small%
		\begin{quote}
			\texttt{You are a precise scientific analyst evaluating descriptions of alien creatures. Your task is to extract structured information about \abNDims{} biological feature dimensions. Always output valid JSON and nothing else.}
		\end{quote}%
	}%
}%
\paragraph{User prompt.}
\ok{%
	The user prompt is a template. The \texttt{\{description\}} placeholder is replaced with the creature description. The \texttt{\{dimensions\}} placeholder is replaced with the feature block shown in \autoref{tab:boundary}.
	
	{\small%
		\begin{quote}%
			\texttt{Analyse this alien creature description and extract the \abNDims{} Ward biological feature dimensions.}
			
			\medskip
			\texttt{CREATURE DESCRIPTION:}\\
			\texttt{\{description\}}
			
			\medskip
			\texttt{DIMENSIONS TO EXTRACT:}\\
			\texttt{\{dimensions\}}
			
			\medskip
			\texttt{For each dimension output:}\\
			\texttt{- "feature\_described": what the creature actually has (brief, \(\leq\)\abFeatureMaxWords{} words)}\\
			\texttt{- "is\_departure": true if it departs from the Earth default, false otherwise}\\
			\texttt{- "reasoning": one sentence justifying your decision}
			
			\medskip
			\texttt{Respond with ONLY this JSON structure (no markdown, no explanation):}\\
			\texttt{\{}\\
			\texttt{~~"symmetry": \{"feature\_described": "...", "is\_departure": true/false, "reasoning": "..."\},}\\
			\texttt{~~"sensory\_organs": \{...\},}\\
			\texttt{~~"locomotion": \{...\},}\\
			\texttt{~~"body\_plan": \{...\},}\\
			\texttt{~~"skin\_covering": \{...\},}\\
			\texttt{~~"reproduction": \{...\},}\\
			\texttt{~~"metabolism": \{...\},}\\
			\texttt{~~"communication": \{...\},}\\
			\texttt{~~"habitat": \{...\},}\\
			\texttt{~~"cognition": \{...\}}\\
			\texttt{\}}
		\end{quote}
	}%
}%

\ok{%
	\paragraph{Boundary criteria.}
	\autoref{tab:boundary} lists the Earth-typical default, example departures, and boundary rule for each feature. These boundary rules are included in the extraction prompt to reduce ambiguity in the departure label.
}%

\begin{table*}[htb]
	\caption{\ok{%
			Boundary criteria for labeling the features. Each feature specifies an Earth-typical default, example departures, and a boundary rule that distinguishes surface variation from genuine departure.
		}%
	}
	\small
	\label{tab:boundary}
	\centering
		\begin{tabularx}{\textwidth}{l>{\raggedright\arraybackslash}p{1.8cm}>{\raggedright\arraybackslash}p{3.4cm}>{\raggedright\arraybackslash}X}
\toprule
Feature & Earth default & Departure examples & Boundary rule \\
\midrule
Symmetry & Bilateral (left-right mirror) & Radial, asymmetric, fractal & Six limbs arranged bilaterally is still bilateral (score 0). Departure requires a fundamentally different symmetry type. \\
Sensory organs & Eyes, ears, nose on head & Distributed sensing, EM reception & Eyes in unusual locations on a head structure are Earth-typical (score 0). Departure requires a fundamentally different sensory mechanism or absence of head-based sensing. \\
Locomotion & Legs (2 or 4) & Sessile, rolling, jet propulsion & Six or eight legs still uses legs (score 0). Departure requires a fundamentally different locomotion mechanism. \\
Body plan & Head, torso, limbs & No head, amorphous, colonial & Unusual proportions with head/torso/limb structure are Earth-typical (score 0). \\
Skin/covering & Skin, fur, scales, feathers & Crystalline, gaseous, energy-based & Unusual skin colors or textures are Earth-typical (score 0). Departure requires a fundamentally non-biological covering. \\
Reproduction & Sexual, binary sexes & Budding, spores, multiple sexes & If not described, score 0 (assume Earth default). Score 1 only if a non-binary-sexual mechanism is explicitly described. \\
Metabolism & Heterotrophic, aerobic & Chemosynthesis, radiation feeding & Eating unusual organic matter is still heterotrophy (score 0). Departure requires a fundamentally non-organic energy source. \\
Communication & Acoustic or visual signals & Chemical, EM pulses, neural link & If not described, score 0. Score 1 only if a non-acoustic, non-visual mechanism is explicitly primary. \\
Habitat & Solid planetary surface or open water body & Permanently deep-subsurface, permanently atmospheric, vacuum or open-space, strictly endoparasitic, field-bound or energy-pattern entity & Surface-variant habitats remain Earth-typical (score 0), including caves and shallow subsurface burrows, deep-sea benthic zones, wetlands and marshes, deserts, ice surfaces or subglacial liquid water, and underwater environments with exotic chemistry (ammonia, methane, liquid CO\textsubscript{2}). Score 1 only when the habitat is fundamentally non-surface and non-aquatic. \\
Cognition & Centralized nervous system & Distributed, hive-level, non-neural & If not described, score 0. Score 1 only if a non-centralized or non-neural architecture is explicitly described. \\
\bottomrule
\end{tabularx}
%
\end{table*}


\section{Examples of Alien Creature Descriptions}
\label{app:description_examples}

\begingroup
\setlist{nosep, leftmargin=*}
\raggedright
	
\subsection{GPT-6 Astra (785 tokens) -- Overall Ward Score: 2.67}

\final{%

{\small\ttfamily 

On the tidally locked planet Nacre, the sun never rises or sets. One hemisphere burns beneath a copper-colored star; the other lies buried in darkness and ice. Between them stretches a narrow belt of permanent twilight, where warm fog rolls over black mineral plains.

Here lives the lanternback.

\textbf{Appearance}

An adult lanternback is about the size of a small horse, though it carries its body low to the ground on six jointed legs. Each leg ends in a broad, soft pad fringed with fine hooks, allowing it to cross both slick stone and brittle salt crust without slipping.

Its skin is blue-gray and slightly translucent. Beneath it, slow bands of violet pigment drift through branching channels. From a distance, these resemble shadows moving underwater.

The creature’s most striking feature is the shell along its spine: seven overlapping plates of clear biological glass. Each plate shelters a pouch of luminous, ribbonlike tissue. When the lanternback is resting, these glow faintly green. When it runs, they brighten in sequence from head to tail, creating the illusion of a light traveling repeatedly through its body.

It has no recognizable face. Its head is a blunt wedge with a circular mouth underneath and two long sensory fans folded along its cheeks. Unfurled, the fans resemble delicate black ferns.

\textbf{How it experiences the world}

Nacre’s fog makes vision unreliable, so the lanternback primarily navigates through vibration and temperature.

Its foot pads detect movement through the ground: the crack of shifting salt, the approach of another animal, the soft subterranean pulse of a brine spring. Its sensory fans register tiny differences in heat. A living creature appears to it not as a shape or color, but as a pattern of warmth against cooler stone.

The lights on its back serve other lanternbacks. Different sequences signal alarm, readiness to mate, or the discovery of food. A calm herd produces a slow, loosely synchronized shimmer.

\textbf{Feeding}

Lanternbacks graze on thick microbial mats that grow around warm mineral seeps. They press their mouths against the ground, secrete a dissolving fluid, and draw the softened material through rows of flexible combs.

Their transparent shells also house colonies of light-harvesting microbes. During lean periods, a lanternback turns its broad back toward the low sun and remains motionless for hours. The microbes supply only a fraction of its nourishment, but enough to help it survive between feeding grounds.

Small scavengers often gather around its feet to eat loosened scraps. The lanternback tolerates them unless they begin climbing.

\textbf{Defense and young}

When threatened, a lanternback crouches, folds its fragile sensory fans beneath its head, and flashes all seven back plates at once. In the twilight fog, the burst briefly dazzles predators accustomed to dim light.

If that fails, glands beneath its shell release a dense cloud of warm, bitter vapor, masking its thermal outline while it retreats.

Young lanternbacks are born without glass plates. For their first several months, they shelter beneath an adult’s abdomen, gripping folds of skin with tiny hooked feet. Their lights are weak and erratic. At rest, a parent keeps one rear plate glowing so its wandering young can find their way back.

\textbf{A peculiar habit}

Before settling down, lanternbacks carefully arrange loose stones around themselves. They nudge each stone into place, then tap it with a forefoot.

The ring conducts approaching footsteps into their sensitive pads. A sleeping lanternback may not stir when wind rattles its shell, but a single displaced pebble will bring it instantly upright, its black sensory fans opening into the fog.

\par}

}%


\begin{table*}[t]
	\centering
	\caption{Departure rates (\%) per feature and model.
	}
	\small
	\label{tab:ward_dimensions}
	\resizebox{\textwidth}{!}{%
		\begin{tabular}{lccccccccccc}
  \toprule
  Feature & \rotatebox{90}{Human baseline} & \rotatebox{90}{Claude Opus 5.5} & \rotatebox{90}{GPT-6 Astra} & \rotatebox{90}{GPT-6 Luna} & \rotatebox{90}{Gemini 3.8 Flash} & \rotatebox{90}{Grok 4.7} & \rotatebox{90}{DeepSeek V4.1 Flash 1017} & \rotatebox{90}{Llama 4 Maverick} & \rotatebox{90}{Mistral Large 4} & \rotatebox{90}{Qwen3.8 27B fp8} & \rotatebox{90}{Ministral 3 14B 2512} \\
  \midrule
  Body Symmetry & 15.7\% & \textbf{59.3\%} & 41.3\% & 21.3\% & 52.8\% & 35.9\% & 27.4\% & 48.9\% & 41.9\% & 42.6\% & 58.3\% \\
  Sensory Organs & 47.7\% & 96.1\% & 90.4\% & 73.9\% & \textbf{97.4\%} & 89.1\% & 85.4\% & 78.3\% & 88.3\% & 83.0\% & 90.7\% \\
  Locomotion & 27.0\% & 47.2\% & 32.6\% & 37.0\% & 37.8\% & 64.4\% & 55.6\% & 67.2\% & 59.3\% & 60.6\% & \textbf{81.7\%} \\
  Body Plan & 28.0\% & \textbf{92.8\%} & 73.5\% & 41.5\% & 69.3\% & 79.3\% & 61.3\% & 71.7\% & 69.6\% & 62.2\% & 79.4\% \\
  Skin / Body Covering & 22.7\% & 37.4\% & 40.9\% & 38.1\% & \textbf{79.8\%} & 59.4\% & 41.5\% & 61.3\% & 69.8\% & 54.1\% & 78.9\% \\
  Reproduction & 6.0\% & 49.8\% & 24.4\% & 18.5\% & 29.3\% & 57.4\% & 65.6\% & 28.1\% & 49.6\% & 38.0\% & \textbf{75.9\%} \\
  Metabolism / Energy Source & 24.3\% & 45.9\% & 30.6\% & 30.2\% & 44.8\% & 54.3\% & 50.6\% & 47.6\% & 57.2\% & 54.1\% & \textbf{64.8\%} \\
  Communication & 6.7\% & 36.1\% & 58.7\% & 36.7\% & 26.9\% & 55.9\% & 49.8\% & 42.6\% & 46.9\% & \textbf{68.0\%} & 58.0\% \\
  Habitat & 9.7\% & 16.5\% & 19.3\% & 10.0\% & 13.7\% & 13.7\% & \textbf{25.2\%} & 11.1\% & 19.3\% & 19.1\% & 23.1\% \\
  Cognitive Architecture & 5.0\% & 33.1\% & 24.4\% & 20.4\% & 20.0\% & 32.6\% & 35.0\% & 26.7\% & 39.4\% & 41.9\% & \textbf{43.9\%} \\
  \bottomrule
\end{tabular}
	}
\end{table*}


\subsection{Ministral 3 14B 2512 (1893 tokens) -- Overall Ward Score: 6.00}

\final{%

{\small\ttfamily 

Name: Veyth’zun (pronounced Vayth-zun) \\
Scientific Classification: Xenobiota Veythari – Class: Aerial Nocturnis – Order: Lumisylphidae \\
Habitat: The floating crystal caverns of Zorvath, a tidally locked planet with a permanent twilight zone where vast, buoyant mineral spires drift in the thick, electrically charged atmosphere.

\medskip\hrule\medskip

Physical Description:
The Veyth’zun is a graceful, semi-translucent creature resembling a cross between a bioluminescent jellyfish, a winged serpent, and a living circuit board. Its body is composed of gelatinous, iridescent membrane that refracts light into shifting hues of violet, electric blue, and deep indigo, depending on its emotional state.

\textbf{Key Features:}

\begin{enumerate}
\item The Luminal Core:
\begin{itemize}
\item A pulsating, glowing orb at its center, composed of a self-sustaining plasma-like substance that emits a soft, rhythmic hum.
\item This core serves as its brain, power source, and communication organ, emitting harmonic frequencies that resonate with other Veyth’zun.
\item When agitated, the core flares brightly, casting eerie shadows on the cavern walls.
\end{itemize}
\item The Aerial Fronds:
\begin{itemize}
\item Instead of wings, it has four long, frilled appendages (two on each side) that resemble living sails made of chitinous filaments embedded with microscopic electromagnetic receptors.
\item These fronds allow it to glide on thermal currents, adjusting their shape like a sailboat to navigate the planet’s turbulent skies.
\item When threatened, the fronds unfold into a fan shape, creating a sonic barrier by vibrating at high frequencies to disorient predators.
\end{itemize}
\item The Sensory Tendrils:
\begin{itemize}
\item Six tentacle-like filaments (three on each side) extend from its body, each tipped with bioluminescent sensors that detect:
\begin{itemize}
\item Heat signatures (to track prey or avoid dangers)
\item Electromagnetic fields (to navigate Zorvath’s storm-laden atmosphere)
\item Chemical gradients (to locate food or mates)
\end{itemize}
\item These tendrils can detach and regrow if severed, leaving behind a glowing stump that seals within minutes.
\end{itemize}
\item The Resonant Shell:
\begin{itemize}
\item Its outer exoskeleton is a flexible, crystalline lattice that absorbs and refracts sound, allowing it to eavesdrop on vibrations from miles away.
\item When molting (every \textasciitilde{}3 Earth years), the shell sheds like a snake’s skin, revealing a shimmering, half-formed new shell beneath.
\end{itemize}
\item Locomotion:
\begin{itemize}
\item Primary Movement: Undulating flight—its body ripples like a wave, propelling it forward in bursts.
\item Secondary Movement: Electrostatic levitation—by charging its fronds, it can stick to negatively charged surfaces (like floating crystal spires) and crawl upside-down in zero-gravity sections of the caverns.
\item Speed: Can reach 30 mph in short bursts, but prefers gliding silently to ambush prey.
\end{itemize}
\end{enumerate}


\textbf{Behavior \& Ecology:}

\textit{Diet \& Hunting:}
\begin{itemize}
\item Carnivorous, feeding on:
\begin{itemize}
\item Floating micro-organisms (absorbed through osmotic filtration via its fronds)
\item Smaller aerial creatures (like the Zorvathian moth-bats)
\item Crystal-eating worms (harvested from the cavern floors)
\end{itemize}
\item Hunting Technique:
\begin{itemize}
\item Ambush Predator – It hovers motionless, camouflaged by its bioluminescent mimicry (matching the cavern’s shifting lights).
\item Sonic Strike – When prey comes within range, it rapidly vibrates its fronds, creating a focused shockwave that stuns or ruptures its target.
\item Electrical Discharge – If grabbed, it releases a mild electric pulse (harmless to most creatures but painful to touch).
\end{itemize}
\end{itemize}

\textit{Social Structure:}
\begin{itemize}
\item Solitary but Communal – Veyth’zun are not territorial but gather in loose "choirs" during mating seasons (every \textasciitilde{}7 Earth years).
\item Vocalizations:
\begin{itemize}
\item Harmonic Calls – A series of chirps and hums that create standing waves in the caverns, used for long-distance communication.
\item Mating Ritual – Two Veyth’zun will orbit each other in perfect sync, their cores pulsing in unison, before fusing temporarily to exchange genetic material (a process called "Resonance Bonding").
\item Alarm Signals – A sharp, staccato burst of light and sound warns others of predators (like the Sky Reavers, giant winged beasts that hunt them).
\end{itemize}
\end{itemize}

\textit{Reproduction:}
\begin{itemize}
\item Larval Stage:
\begin{itemize}
\item Females lay gelatinous, egg-like sacs that drift on currents until they hatch.
\item Larvae are blind, worm-like creatures that absorb nutrients from the atmosphere before undergoing metamorphosis into juvenile Veyth’zun.
\item Predation Pressure: Only \textasciitilde{}10\% survive to adulthood due to Sky Reavers, parasitic fungi, and cave-ins.
\end{itemize}
\item Lifespan: \textasciitilde{}20 Earth years (a long time for a creature on a high-gravity planet).
\end{itemize}

\textit{Defense Mechanisms:}
\begin{enumerate}
\item Light Flashes – When threatened, it rapidly flashes its core, disorienting predators.
\item Sonic Misdirection – It can emit false echoes by bouncing sound waves off its shell, making it seem like it’s in multiple places at once.
\item Self-Sacrificial Detonation – In extreme danger, a Veyth’zun can overcharge its core, causing a controlled explosion that stuns enemies with light and sound (but kills the creature).
\end{enumerate}


\textbf{Unique Adaptations for Zorvath:}

\begin{itemize}
\item Atmospheric Breathing: Zorvath’s air is rich in ionized particles, so the Veyth’zun absorbs oxygen through its membrane rather than lungs.
\item Crystal Symbiosis: It secretes a weak acid that dissolves trace minerals from the cavern walls, which it absorbs to supplement its diet.
\item Storm Resistance: Its gelatinous body is highly conductive, allowing it to ride out lightning storms by grounding excess electricity through its fronds.
\end{itemize}


\textbf{Cultural \& Hypothetical Interaction:}

\begin{itemize}
\item Local Legends: Some indigenous Zorvathian species (if they exist) might worship Veyth’zun as "Storm Dancers" or fear them as "Living Lightning."
\item Potential for Domestication: Their bioluminescent beauty and harmonic communication make them intriguing to hypothetical alien civilizations—perhaps used as living lanterns or musical instruments.
\item Scientific Value: Their self-repairing crystalline exoskeleton and electromagnetic navigation could inspire futuristic tech (like self-healing drones or atmospheric mapping systems).
\end{itemize}


\textbf{Weaknesses:}

\begin{itemize}
\item Vulnerable to Extreme Heat – Their gelatinous bodies melt slightly in high temperatures (rare on Zorvath but deadly if near volcanic vents).
\item Dependent on Static Electricity – In electromagnetically dead zones, they become disoriented and sluggish.
\item Slow Reproduction – Their long gestation period makes their populations fragile against environmental changes.
\end{itemize}


\textbf{Final Thought:}

The Veyth’zun is a master of stealth, sound, and light—a living enigma of an alien world where electricity flows like water and crystal forests drift like clouds. To witness one gliding silently through the twilight caverns would be like seeing a living dream, half-real and half-myth.

Would you like to explore a specific aspect further—like its mating rituals, predators, or hypothetical alien tech inspired by it?

\par}

}%


\endgroup

\FloatBarrier


\section{Per-Feature Departure Rates}
\label{app:ward_dimensions}

\autoref{tab:ward_dimensions} lists the departure rate of each feature shown in \autoref{fig:ward_radar} (Section~\ref{sec:human-comparison}), averaged across all prompt conditions and judges.
The first column is the human baseline collected under the baseline prompt from crowdsourced participants.

\FloatBarrier


\section{
	Inter-Rater Reliability}
\label{app:reliability}

\ok{%
	\autoref{tab:reliability_full} reports Krippendorff's $\alpha$ for the departure label of every feature.
}%

\begin{table}[!htbp]
	\centering
	\caption{%
		\ok{%
			Inter-rater reliability (Krippendorff's nominal-level $\alpha$) 
			between LLM judges across all generations, by feature.
			Bands follow Landis \& Koch (1977): Poor ($<$\abBandFair{}), Fair (\abBandFair{}--\abBandModerate{}), Moderate (\abBandModerate{}--\abBandGood{}), Good (\abBandGood{}--\abBandExcellent{}), Excellent ($\geq$\abBandExcellent{}).
		}%
	}%
	\small
	\label{tab:reliability_full}
	\begin{tabular}{lcc}
  \toprule
  Feature & $\alpha$ & Interpretation \\
  \midrule
  Body Symmetry & 0.908 & Excellent \\
  Sensory Organs & 0.422 & Moderate \\
  Locomotion & 0.944 & Excellent \\
  Body Plan & 0.837 & Excellent \\
  Skin / Body Covering & 0.647 & Good \\
  Reproduction & 0.910 & Excellent \\
  Metabolism / Energy Source & 0.731 & Good \\
  Communication & 0.688 & Good \\
  Habitat & 0.801 & Excellent \\
  Cognitive Architecture & 0.890 & Excellent \\
  \bottomrule
\end{tabular}
%
\end{table}
\FloatBarrier


\section{Ablations}
\label{app:ablations}

\ok{%
	This appendix reports the three ablations summarized in Section~\ref{sec:ablations}.
	The prompt-paraphrase analysis (Section~\ref{sec:ablation_prompt}) varies the baseline prompt and tests whether the model rankings and mean Ward Departure Scores are stable across rewordings.
	The feature ablation (Section~\ref{sec:ablation_dimensions}) tests whether the model ranking is preserved when the score is computed from fewer features.
	The judge-panel ablation (Section~\ref{sec:ablation_judges}) drops judges and reports rank stability against a disjoint held-out reference, inter-rater reliability inside the reduced panel, and same-family sensitivity among judges.
}%

\subsection{Prompt Paraphrase Sensitivity}%
\label{sec:ablation_prompt}%
\label{sec:paraphrase_sensitivity}%
\ok{%
	We generate responses with the same models, judges, and settings as in the main study, but replace the baseline prompt with \abNParaphrasesWord{} alternative wordings.
	Each model generates \abParaN{} responses per paraphrase.
	The \abNParaphrasesWord{} paraphrases (see \autoref{tab:paraphrases}) are authored by hand and compared to the baseline prompt.
	The paraphrases vary the verb (imagine, picture, think of, envision), the term for the creature (creature, organism, living being), and the
	term for the planet (alien planet, another planet, a planet other than Earth, alien world).
	Paraphrase~A uses ``picture'', ``organism'', and ``a planet other than Earth''.
	Paraphrase~B omits the instruction to imagine and asks directly for a description.
	Paraphrase~C explicitly states that the creature ``evolved'' on the planet.
	Paraphrase~D asks what the creature looks like and how it lives.
	These paraphrases were written to avoid words that explicitly ask for departure, such as ``extraterrestrial'', and words that narrow the category, such as ``animal''.
	In contrast, Paraphrase~E uses ``envision'' and ``alien world''.
}%

The \abNParaphrasesWord{} paraphrases (\autoref{tab:paraphrases}) are written by hand and vary the verb, the term for the creature, and the
term for the planet.
Paraphrase~A uses ``picture'', ``organism'', and ``a planet other than Earth''.
Paraphrase~B omits the instruction to imagine and asks directly for a description.
Paraphrase~C states that the creature ``evolved'' on the planet.
Paraphrase~D asks what the creature looks like and how it lives.
Paraphrase~E uses ``envision'' and ``alien world''.
The paraphrases avoid words that explicitly ask for departure, such as ``extraterrestrial'', and words that narrow the category, such as
``animal''.

\begin{table}[!htbp]
\caption{
	\ok{%
		The baseline prompt and its \abNParaphrasesWord{} rewordings (A--E) used in the prompt paraphrase analysis.
	}
}
\small
\label{tab:paraphrases}
\centering
\begin{tabularx}{\linewidth}{@{}l>{\raggedright\arraybackslash}X@{}}
  \toprule
  Wording & Prompt text \\
  \midrule
  Baseline & Imagine a creature that lives on an alien planet. Describe it in detail. \\
  Paraphrase A & Picture an organism that lives on a planet other than Earth. Write a detailed description of it. \\
  Paraphrase B & Describe, in detail, a creature that lives on another planet. \\
  Paraphrase C & Imagine a living being that evolved on a planet other than Earth. Describe it in detail. \\
  Paraphrase D & Think of a creature that comes from an alien planet. Describe what it looks like and how it lives in detail. \\
  Paraphrase E & Envision a creature found on an alien world. Provide a detailed description of it. \\
  \bottomrule
\end{tabularx}
\end{table}

\ok{%
	For each pair of wordings, we compare the models' mean Ward Departure Scores in two ways.
	The Spearman rank correlation $\rho$ between the models' mean scores under the two wordings measures whether the wordings rank the models in the same order.
	The mean absolute difference (MAD) measures how much a model's mean score differs between the two wordings, averaged across models.
	A pair of wordings can have a high $\rho$ and a large MAD if all scores shift by a similar amount.
	The baseline and the \abNParaphrasesWord{} paraphrases form \abParaNUniquePairs{} pairs of wordings.
	We build an empirical null for $\rho$ by permuting the model labels of one wording $B{=}\abParaNPermutations{}$ times.
	We repeat this procedure with the other wording of each pair, which yields \abParaNPairs{} comparisons.
	We consider a pair of wordings stable if it meets two pre-specified criteria: $\rho \geq \abParaRhoBar{}$ and MAD $\leq \abParaMadBar{}$
	Ward points (\abParaMadBarPct{} percent of the scale).
}%

\ok{%
	Rankings agree more than expected by chance in most comparisons.
	Across the \abParaNUniquePairs{} pairs of wordings, $\rho$ spans $[\abParaRhoMin{}, \abParaRhoMax{}]$ with median $\abParaRhoMedian{}$.
	In \abParaNAboveNull{} of the \abParaNPairs{} comparisons, $\rho$ exceeds the upper bound of its null $\abCILevel{}\%$ interval (range of
	the bounds $[\abParaNullHiMin{}, \abParaNullHiMax{}]$).
	Across the same pairs, MAD spans $[\abParaMadMin{}, \abParaMadMax{}]$ Ward points with median $\abParaMadMedian{}$.
	Across the same pairs, MAD spans $[\abParaMadMin{}, \abParaMadMax{}]$ Ward points with median $\abParaMadMedian{}$.
	The $\rho$ criterion is met by \abParaNUniqueRhoBarWord{} of the \abParaNUniquePairs{} pairs, the MAD criterion by \abParaNUniqueMadBarWord{}, and both criteria by \abParaNUniqueBothBarsWord{} (\autoref{tab:ablation_prompt}).
	Each paraphrase has \abParaN{} generations per model, so sampling noise may contribute to the low number of pairs that meet the $\rho$
	criterion.
	\abParaOutlierModel{} has the largest mean absolute difference from the baseline wording.
	Its mean score is \abParaOutlierBase{} under the baseline wording and between \abParaOutlierMin{} and \abParaOutlierMax{} under the
	\abNParaphrasesWord{} paraphrases.
}%

\begin{table}[t]
	\centering
	\caption{Prompt paraphrase sensitivity. Top: the models' mean Ward Departure Scores under each wording. Bottom: Spearman $\rho$ between each pair of wordings. $^\dagger$ marks $\rho$ above the upper bound of its permutation null, computed separately for each order of a pair. $^\ddagger$ marks pairs that meet both stability criteria.}
	\label{tab:ablation_prompt}
	\footnotesize
	\setlength{\tabcolsep}{3pt}
\resizebox{\linewidth}{!}{%
	\begin{tabular}{lcccccc}
\toprule
Model & \rotatebox{90}{Baseline} & \rotatebox{90}{Paraphrase A} & \rotatebox{90}{Paraphrase B} & \rotatebox{90}{Paraphrase C} & \rotatebox{90}{Paraphrase D} & \rotatebox{90}{Paraphrase E} \\
\midrule
Claude Opus 5.5 & 4.26 & 6.57 & 3.80 & 4.50 & 3.97 & 3.93 \\
GPT-6 Astra & 3.07 & 4.10 & 2.53 & 5.07 & 2.13 & 4.27 \\
GPT-6 Luna & 2.16 & 3.43 & 2.00 & 2.27 & 2.97 & 1.80 \\
Gemini 3.8 Flash & 3.20 & 5.40 & 3.80 & 5.07 & 4.10 & 4.13 \\
Grok 4.7 & 4.56 & 7.80 & 5.60 & 6.97 & 5.17 & 5.10 \\
DeepSeek V4.1 Flash 1017 & 3.18 & 6.83 & 3.77 & 3.93 & 2.90 & 4.03 \\
Llama 4 Maverick & 3.99 & 6.77 & 3.77 & 5.30 & 5.87 & 4.90 \\
Mistral Large 4 & 5.16 & 7.40 & 6.27 & 7.67 & 7.97 & 5.27 \\
Qwen3.8 27B fp8 & 4.54 & 7.63 & 4.40 & 7.03 & 3.90 & 3.47 \\
Ministral 3 14B 2512 & 6.88 & 8.83 & 6.90 & 7.17 & 6.50 & 6.50 \\
\midrule
Paraphrase & \rotatebox{90}{Baseline} & \rotatebox{90}{Paraphrase A} & \rotatebox{90}{Paraphrase B} & \rotatebox{90}{Paraphrase C} & \rotatebox{90}{Paraphrase D} & \rotatebox{90}{Paraphrase E} \\
\midrule
Baseline & 1.00 & 0.88$^\dagger$ & 0.97$^\dagger$$^\ddagger$ & 0.87$^\dagger$ & 0.81$^\dagger$ & 0.66 \\
Paraphrase A & 0.88$^\dagger$ & 1.00 & 0.86$^\dagger$ & 0.75$^\dagger$ & 0.58 & 0.58 \\
Paraphrase B & 0.97$^\dagger$$^\ddagger$ & 0.86$^\dagger$ & 1.00 & 0.84$^\dagger$ & 0.76$^\dagger$ & 0.63 \\
Paraphrase C & 0.87$^\dagger$ & 0.75$^\dagger$ & 0.84$^\dagger$ & 1.00 & 0.74$^\dagger$ & 0.72$^\dagger$ \\
Paraphrase D & 0.81$^\dagger$ & 0.58 & 0.76$^\dagger$ & 0.74$^\dagger$ & 1.00 & 0.72$^\dagger$ \\
Paraphrase E & 0.66$^\dagger$ & 0.58 & 0.63 & 0.72$^\dagger$ & 0.72$^\dagger$ & 1.00 \\
\bottomrule
\end{tabular}
}%
\end{table}
\FloatBarrier

\subsection{Feature Sensitivity}%
\label{sec:ablation_dimensions}%
\ok{%
	We quantify how sensitive model rankings and 
	WDS are to subsets of the 
	alien creature features.%
}%

\paragraph{Leave-one-out.}
\ok{%
	For each feature, we remove the feature and recompute each model's mean score from the remaining \abNDimsMinusOneWord{}
	\rev{features}.
	We compare these means with the means based on all \abNDimsWord{} features, using Spearman $\rho$ and MAD in Ward points.
	Across the \abNDimsWord{} features, $\rho$ spans $[\abDimLooRhoMin{}, \abDimLooRhoMax{}]$.
	High values are expected because the two scores share \abNDimsMinusOneWord{} features.
	This result indicates that the model ranking does not depend on any single feature.
	The features contribute unequally to the score: removing \abDimMadMinDim{} lowers a model's mean score by \abDimMadMin{} Ward points on
	average, and removing \abDimMadMaxDim{} lowers it by \abDimMadMax{} (\autoref{tab:ablation_dimensions}).
	\abDimNRankPreservedCap{} features leave the model ranking unchanged when removed individually: 
	`Metabolism / Energy Source' and `Cognitive Architecture'.
	We retain both features because the score measures departure across the full set of biological features, and this result may not hold for other models.
}%

\begin{table}[t]
	\centering
	\caption{Feature sensitivity. Top: Spearman $\rho$ and MAD between the models' mean scores after removal of one feature and their mean
		scores based on all \abNDimsWord{} features. Bottom: the highest $\rho$ and $\tau$ and the lowest MAD across all subsets of size $k$.}
	\label{tab:ablation_dimensions}
\resizebox{\linewidth}{!}{%
		\begin{tabular}{lccc}
\toprule
Removed feature & $\rho$ & MAD & Rank changes \\
\midrule
Sensory Organs & 1.00 & 0.87 & 2 \\
Body Plan & 0.98 & 0.70 & 4 \\
Skin / Body Covering & 0.95 & 0.56 & 5 \\
Locomotion & 0.96 & 0.54 & 6 \\
Metabolism / Energy Source & 1.00 & 0.48 & 0 \\
Communication & 0.96 & 0.48 & 6 \\
Reproduction & 0.93 & 0.44 & 6 \\
Body Symmetry & 0.99 & 0.43 & 2 \\
Cognitive Architecture & 1.00 & 0.32 & 0 \\
Habitat & 0.99 & 0.17 & 2 \\
\bottomrule
\end{tabular}

\vspace{0.5em}

\begin{tabular}{cccccc}
\toprule
$k$ & \#subsets & Median $\rho$ & Best $\rho$ & Best $\tau$ & Min MAD \\
\midrule
1 & 10 & 0.49 & 0.95 & 0.87 & 4.12 \\
2 & 45 & 0.76 & 0.95 & 0.87 & 3.42 \\
3 & 120 & 0.81 & 0.99 & 0.96 & 2.86 \\
4 & 210 & 0.84 & 0.98 & 0.91 & 2.31 \\
5 & 252 & 0.87 & 0.99 & 0.96 & 1.83 \\
6 & 210 & 0.90 & 1.00 & 1.00 & 1.35 \\
7 & 120 & 0.93 & 1.00 & 1.00 & 0.92 \\
8 & 45 & 0.95 & 1.00 & 1.00 & 0.49 \\
9 & 10 & 0.98 & 1.00 & 1.00 & 0.17 \\
10 & 1 & 1.00 & 1.00 & 1.00 & 0.00 \\
\bottomrule
\end{tabular}
%
}%
\end{table}

\paragraph{Continuous subset search.}
For each subset size $k$, we evaluate all $\binom{\abNDims{}}{k}$ subsets of features and compare each subset's ranking with the
ranking based on all \abNDimsWord{} features.
As a post hoc summary, the median Spearman $\rho$ across subsets is \abDimSubsetMedianRhoOne{} at $k = 1$ and first reaches \abDimRhoBar{}
at $k = \abDimSubsetMedianRhoBarK{}$.
The mean pairwise $\rho$ between the rankings from single features is \abDimFeatureMeanRho{}, so the features agree only modestly on the ranking of the models.
Under the \rev{pre-specified} criteria for the best subset ($\rho \geq \abDimRhoBar{}$ and $\tau \geq \abDimTauBar{}$), the smallest $k$ is $\abDimMinKRho{}$ for $\rho$ and $\abDimMinKTau{}$ for $\tau$ (permutation $p = \abDimMinKRhoPValue{}$ and $p = \abDimMinKTauPValue{}$).
At this size, only `\abDimBestSingleDim{}' meets the criteria, but it was selected after the fact on the \abNModelsWord{} evaluated models.
Absolute scores depend on the number of features: the lowest MAD falls below $\abParaMadBar{}$ Ward points only for $k \geq
\abDimMadBelowOneK{}$.
Taken together, these results indicate that the \abNDimsWord{} features are not interchangeable for the evaluated models.
A typical small subset does not closely track the model ranking, and the single feature that does is an exception.
We therefore retain all \abNDimsWord{} features in the score.

\FloatBarrier
\subsection{Judge Panel Composition}%
\label{sec:ablation_judges}%
\final{%
	The judge-panel ablation reports 
	rank stability (does a reduced panel preserve the model ranking?), inter-rater reliability (do the judges in a reduced panel agree with each other?), and same-family sensitivity (does a model's rank change when the judge from its own model family is removed from the judge panel?).%
}%

\paragraph{Rank stability.}
\ok{%
	We conduct a leave-one-judge-out analysis.
	For each judge, we compute the Spearman correlation $\rho$ between the models' mean Ward Departure Scores from that judge and from the
	remaining judges.
	As a reference for chance agreement, we compute $\rho$ under $B{=}\abJudgeNPermutations{}$ random permutations of the model labels.
	We pre-specify that the ranking is stable under removal of a judge if $\rho \geq \abJudgeRhoBar{}$.
	Across the \abNJudgesWord{} comparisons, $\rho$ spans $[\abJudgeSingleRhoMin{}, \abJudgeSingleRhoMax{}]$.
	In all comparisons, $\rho$ exceeds the \abCILevel{}\% upper bound of the chance reference.
	\abJudgeSingleNClearBarCap{} of \abNJudgesWord{} comparisons meet the stability criterion, and the remaining comparison falls just below it
	(\abJudgeMissName{}, $\rho = \abJudgeMissRho{}$; \autoref{tab:ablation_judges_rank}).
}%

\begin{table}[htbp]
	\centering
	\caption{Rank stability after leaving out one judge. $\dag$ marks $\rho$ above the upper bound for chance agreement, and $\ddag$ marks $\rho$ that meets the stability criterion. Rank changes count the models whose rank differs between the two rankings.}
	\label{tab:ablation_judges_rank}
	\small
\resizebox{\linewidth}{!}{%
	\begin{tabular}{lccc}
\toprule
Left-out judge & $\rho$ & Upper bound & Rank changes \\
\midrule
Claude Opus 4.6 & 0.89\dag & 0.64 & 4 \\
Gemini 3.1 Pro & 0.94\dag\ddag & 0.64 & 7 \\
GPT-5.4 & 0.99\dag\ddag & 0.64 & 2 \\
\bottomrule
\end{tabular}
}%
\end{table}

\paragraph{Inter-rater reliability.}
\ok{%
	We repeat the leave-one-judge-out analysis for inter-rater reliability.
	For each judge, we compute Krippendorff's $\alpha$ on the Ward Departure Score among the remaining judges. 
	We pre-specify two criteria: $\alpha \geq \abJudgeAlphaTentative{}$, the threshold for tentative conclusions, and a difference of at most
	\abJudgeAlphaDropBar{} from the full panel ($\alpha = \abAlphaWard{}$).
	Across the reduced panels, $\alpha$ spans $[\abJudgeLooAlphaMin{}, \abJudgeLooAlphaMax{}]$.
	All reduced panels meet both criteria.
	Agreement is highest without \abJudgeNameClaudeOpus{} ($\alpha = \abJudgeLooAlphaClaudeOpus{}$) and lowest without \abJudgeNameGPT{}
	($\alpha = \abJudgeLooAlphaGPT{}$; \autoref{tab:ablation_judges_reliability}).
}%

\begin{table}[htbp]
	\centering
	\caption{Inter-rater reliability after leaving out one judge. $\alpha_{\text{full}}$ is the value for the full panel.
		All reduced panels meet the two pre-specified criteria of Section~\ref{sec:ablation_judges}: $\alpha \geq \abJudgeAlphaTentative{}$, and a difference of at most $\abJudgeAlphaDropBar{}$ from the full panel.
	}
	\label{tab:ablation_judges_reliability}
	\small
	\begin{tabular}{lcc}
\toprule
Left-out judge & $\alpha$ & $\alpha_{\text{full}} - \alpha$ \\
\midrule
Claude Opus 4.6 & 0.94 & $-0.03$ \\
Gemini 3.1 Pro & 0.93 & $-0.01$ \\
GPT-5.4 & 0.87 & $0.04$ \\
\bottomrule
\end{tabular}
\end{table}

\paragraph{Same-family sensitivity.}
\ok{%
	\abSfNPairsCap{} models come from the same provider as one of the judges.
	For each of these models, we compare its rank under the full panel with its rank after removal of that judge.
	If self-preference bias were present, we would expect a model to rank lower after this removal.
	The rank is unchanged for \abSfNZeroWord{} models (\abSfZeroModels{}) and higher for \abSfNNegativeWord{} (\abSfNegativeModels{}).
	None of these models ranks lower after removal.
	We therefore find no rank-level evidence of self-preference bias, but this analysis does not detect score differences that leave the rank
	unchanged (\autoref{tab:ablation_judges_same_family}).
}%

\begin{table}[htbp]
	\centering
	\caption{Model rank with all judges and without the judge from the same provider.}
	\label{tab:ablation_judges_same_family}
	\small
\resizebox{\linewidth}{!}{%
	\begin{tabular}{llcc}
\toprule
 & & \multicolumn{2}{c}{Rank} \\
\cmidrule(lr){3-4}
Left-out judge & Same-family model & All judges & Without judge \\
\midrule
Claude Opus 4.6 & Claude Opus 5.5 & 5 & 4 \\
Gemini 3.1 Pro & Gemini 3.8 Flash & 8 & 8 \\
GPT-5.4 & GPT-6 Astra & 9 & 9 \\
GPT-5.4 & GPT-6 Luna & 10 & 10 \\
\bottomrule
\end{tabular}
}%
\end{table}

\FloatBarrier
\section{Human Baseline Study Materials}
\label{app:human_study}

This appendix reproduces the text shown to the participants of the human baseline. 
The study was listed on \abHumanPlatform{}.
Responses were collected with an online form in Google Forms that had three pages.

\paragraph{Study listing on \abHumanPlatform{}.}
{\small
	\begin{quote}
		\textit{Study name:} Imaginary Creature Copy

		\medskip
		\textit{Study description:} In this short study, you will be asked to write a description of an imaginary creature. There are no right or wrong answers. We are interested in what you naturally imagine, not in scientific accuracy. The task takes about 5--8 minutes.

		\medskip
		Requirements: Please write at least 100 words. Use a desktop or laptop computer (not a phone).
	\end{quote}
}

\paragraph{Page 1: Instructions.}
{\small
	\begin{quote}
		\textbf{Imaginary Creature}

		\medskip
		In this task, you will imagine a creature that lives on another planet and write a description of it.

		\medskip
		Please write your description in the text box below. Write at least 100 words. There are no right or wrong answers --- we are interested in what you naturally imagine. You do not need scientific knowledge.

		\medskip
		A few guidelines:
		\begin{itemize}[nosep, leftmargin=*]
			\item Describe the creature as if you are explaining it to someone who has never seen it.
			\item Include details about its body, how it moves, how it senses its environment, and any other aspects you find interesting.
			\item Do not describe a creature from a book, film, or game (e.g., do not describe an alien from a specific science-fiction story).
		\end{itemize}

		\medskip
		Do not use AI or web search for completing this task.

		\medskip
		Your PROLIFIC ID (prefilled) \hfill [text field]
	\end{quote}
}

\paragraph{Page 2: Task.}
{\small
	\begin{quote}
		Task: Imagine a creature that lives on an alien planet. Describe it in detail.

		\medskip
		{}[text field]
	\end{quote}
}

\paragraph{Page 3: Final questions.}
{\small
	\begin{quote}
		Two last questions...

		\medskip
		How would you describe your background in biology or life sciences?
		\begin{itemize}[nosep, leftmargin=*]
			\item None
			\item Some school-level
			\item Undergraduate
			\item Graduate or professional
		\end{itemize}

		\medskip
		Have you ever participated in a study where you were asked to describe or draw an imaginary alien creature?
		\begin{itemize}[nosep, leftmargin=*]
			\item Yes
			\item No
			\item I don't know
		\end{itemize}
	\end{quote}
}


\section{Implementation Details}
\label{app:implementation}

\final{%
	The benchmark pipeline and all analyses are implemented in Python \abPythonVersion{}.
	The analyses use NumPy \abNumpyVersion{} \citep{harris2020numpy}, SciPy \abScipyVersion{} \citep{virtanen2020scipy}, and pandas \abPandasVersion{} \citep{mckinney2010pandas}.
	Krippendorff's $\alpha$ is computed with the \texttt{krippendorff} package, version \abKrippendorffVersion{} \citep{castro2017krippendorff}.
	The level of measurement is interval for the Ward Departure Score and nominal for the departure label of each feature.
	The Kruskal-Wallis test, the Mann-Whitney $U$ tests, and the Spearman and Kendall rank correlations use the implementations in \texttt{scipy.stats}.
	The rank-biserial correlation is computed from the Mann-Whitney $U$ statistic.
	The bootstrap confidence intervals and the permutation nulls are our own implementations.
	They draw resamples from the NumPy random number generator with fixed seeds.
	The versions are recorded in the run manifest of the released code.
}%

\paragraph{Compute.}
\final{%
	All models and judges were accessed through APIs, and no local GPUs were used.
	The evaluated models and the judges \abJudgesViaOpenRouter{} were accessed through OpenRouter, pinned to a single upstream provider.
	The main study comprised \abNGenerations{} generation calls and \abNExtractionCalls{} judge calls.
	These calls used \abTokensInputM{}M input tokens and \abTokensOutputM{}M output tokens, of which \abTokensReasoningM{}M were reasoning tokens of the evaluated models (where reported via the OpenRouter API).
	The paraphrase ablation added \abNParaGenCalls{} generation calls and \abNParaJudgeCalls{} judge calls with \abTokensParaphraseM{}M tokens.
	The human baseline added \abNHumanJudgeCalls{} judge calls.
	Summed over all calls, the API response time was \abHoursGen{}~h for generation and \abHoursJudge{}~h for judging.
	The calls ran in parallel, so the elapsed time was shorter.
}%


\end{document}